\documentclass[10pt,twocolumn,letterpaper]{article}

\usepackage{cvpr}              
\usepackage[accsupp]{axessibility}
\usepackage{graphicx}
\usepackage{amsmath}
\usepackage{amssymb}
\usepackage{booktabs}
\usepackage[table]{xcolor}
\usepackage{color}
\usepackage{times}
\usepackage{epsfig}
\usepackage{amsbsy}
\usepackage{xcolor}
\usepackage{xspace}
\usepackage{algorithm}
\usepackage{algorithmic}
\usepackage{multirow}
\usepackage{multicol}
\usepackage{float}
\usepackage{relsize}
\usepackage{soul}
\usepackage{tabularx}
\usepackage{makecell}
\usepackage{bbm}
\usepackage{wrapfig,lipsum,booktabs}

\usepackage[pagebackref,breaklinks,colorlinks]{hyperref}
\DeclareMathOperator*{\argmax}{arg\, max}  

\usepackage[capitalize]{cleveref}
\crefname{section}{Sec.}{Secs.}
\Crefname{section}{Section}{Sections}
\Crefname{table}{Table}{Tables}
\crefname{table}{Tab.}{Tabs.}
\definecolor{aliceblue}{rgb}{0.94, 0.97, 1.0}

\def\confName{CVPR}
\def\confYear{2023}

\begin{document}

\title{C-SFDA: A Curriculum Learning Aided Self-Training Framework for Efficient Source Free Domain Adaptation}




\author{
Nazmul Karim$^{*}$,
Niluthpol Chowdhury Mithun$^{\dag}$,
Abhinav Rajvanshi$^{\dag}$,  Han-pang Chiu$^{\dag}$ \\ Supun Samarasekera$^{\dag}$,  Nazanin Rahnavard$^{*}$ \and
{\normalsize $^{*}$Department of ECE, UCF, Orlando FL USA} \and {\normalsize $^{\dag}$SRI International, Princeton, NJ, USA} 
\and
\texttt{\small $^{*}$nazmul.karim18@knights.ucf.edu} \and \texttt{\small $^{\dag}$firstname.lastname@sri.com}
}

\maketitle

{ \renewcommand{\thefootnote}%
    {\fnsymbol{footnote}}
  \footnotetext[1]{Most of this work was done during Nazmul Karim's internship with SRI International. Project Page: \url{https://sites.google.com/view/csfdacvpr2023/home}}
}

\begin{abstract}

Unsupervised domain adaptation (UDA) approaches focus on adapting models trained on a labeled source domain to an unlabeled target domain. 
In contrast to UDA, source-free domain adaptation (SFDA) is a more practical setup as access to source data is no longer required during adaptation. Recent state-of-the-art (SOTA) methods on SFDA mostly focus on pseudo-label refinement based self-training which generally suffers from two issues: i) inevitable occurrence of noisy pseudo-labels that could lead to early training time memorization, ii) refinement process requires maintaining a memory bank which creates a significant burden in resource constraint scenarios. To address these concerns, we propose C-SFDA, a curriculum learning aided self-training framework for SFDA that adapts efficiently and reliably to changes across domains based on selective pseudo-labeling. Specifically, we employ a curriculum learning scheme to promote learning from a restricted amount of pseudo labels selected based on their reliabilities. This simple yet effective step successfully prevents label noise propagation during different stages of adaptation and eliminates the need for costly memory-bank based label refinement. Our extensive experimental evaluations on both image recognition and semantic segmentation tasks confirm the effectiveness of our method. C-SFDA is also applicable to online test-time domain adaptation and outperforms previous SOTA methods in this task.
\end{abstract}

\section{Introduction}
\label{sec:intro}
\vspace{-1mm}
Deep neural network (DNN) models have achieved remarkable success in various visual recognition tasks~\cite{liang2020polytransform,he2016deep,dosovitskiy2020image,liu2022adaptive}. However, even very large DNN models often suffer significant performance degradation when there is a distribution or domain shift~\cite{xia2020structure, peng2019moment} between training (source) and test (target) domains. To address the problem of domain shifts, various Unsupervised Domain Adaptation (UDA)~\cite{RSDA,CAN} algorithms have been developed over recent years. Most UDA techniques require access to labeled source domain data during adaptation, which limits their application in many real-world scenarios, \eg source data is private, or adaptation in edge devices with limited computational capacity. In this regard, source-free domain adaptation setting has recently gained significant interest~\cite{universalSFDA,kundu2020towards, yang2020unsupervised}, which considers the availability of only source pre-trained model and unlabeled target domain data.

\begin{figure*}[htb]
\centering
\includegraphics[width=0.95\linewidth]{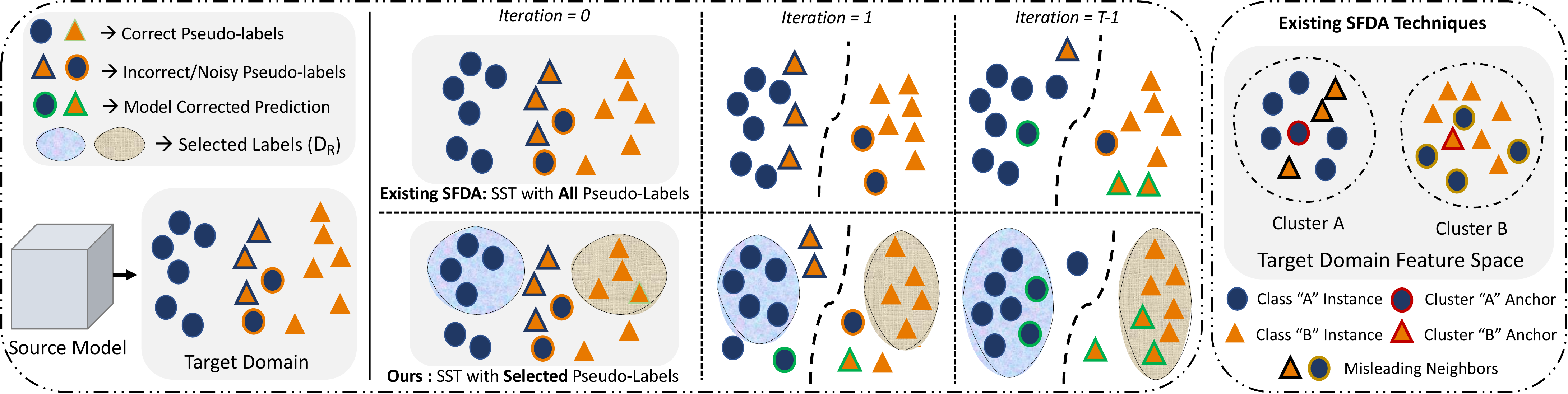}
\vspace{-2mm}
\caption{\footnotesize \textbf{\emph{Left:}} In SFDA, we only have a source model that needs to be adapted to the target data. Among the source-generated pseudo-labels, a large portion is noisy which is important to avoid during supervised self-training (SST) with regular cross-entropy loss. Instead of using all pseudo-labels, we choose the most reliable ones and effectively propagate high-quality label information to unreliable samples. As the training progresses, the proposed selection strategy tends to choose more samples for SST due to the improved average reliability of pseudo-labels. Such a restricted self-training strategy creates a model with better discrimination ability and eventually corrects the noisy predictions. Here, $T$ is the total number of iterations. \textbf{\emph{Right:}} While existing SFDA techniques leverages cluster structure knowledge in the feature space, there may exist many misleading neighbors\textbf{---} pseudo-labels of neighbors' that are different from the anchors' true label. Therefore, clustering-based label propagation inevitably suffers from label noise in subsequent training.}
\label{fig:main}
\vspace{-3.5mm}
\end{figure*}

Recent state-of-the-art SFDA methods (e.g., SHOT~\cite{SHOT}, NRC~\cite{yang2021exploiting}, G-SFDA~\cite{G-SFDA}, AdaContrast~\cite{chen2022contrastive}) mostly rely on the self-training mechanism that is guided by the source pre-trained model generated pseudo-labels (PLs). Label refinement using the knowledge of per-class cluster structure in feature space is recurrently used in these methods. At early stages of adaptation, the label information formulated based on cluster structure can be severely misleading or noisy; shown in Fig.~\ref{fig:main}. As the adaptation progresses, this label noise can negatively impact the subsequent cluster structure as the key to learning meaningful clusters hinges on the quality of pseudo-labels itself. Therefore, the inevitable presence of label noise at early training time is a critical issue in SFDA and requires proper attention. Furthermore, distributing cluster knowledge among neighbor samples requires a memory bank~\cite{chen2022contrastive, SHOT} which creates a significant burden in resource-constraint scenarios. In addition, most memory bank dependent SFDA techniques are not suitable for online test-time domain adaptation~\cite{wang2021tent,wang2022continual}; an emerging area of UDA that has gained traction in recent times.
\emph{Designing a memory-bank-free SFDA approach that can guide the self-training with highly precise pseudo-labels is a very challenging task and a major focus of this work.}

In our work, we focus on increasing the reliability of generated pseudo-labels without using a memory-bank and clustering-based pseudo-label refinement. Our analysis shows that avoiding early training-time memorization (ETM) of noisy labels encourages noise-free learning in subsequent stages of adaptation. We further analyze that even with an expensive label refinement technique in place, learning equally from all labels eventually leads to label-noise memorization. Therefore, we employ a curriculum learning-aided self-training framework, C-SFDA, that prioritizes learning from easy-to-learn samples first and hard samples later on. We show that one can effectively identify the group of easy samples by utilizing the reliability of pseudo-labels, \ie prediction confidence and uncertainty. We then follow a carefully designed curriculum learning pipeline to learn from highly reliable (easy) samples first and gradually propagate more refined label information among less reliable (hard) samples later on. In addition to the self-training, we facilitate unsupervised contrastive representation learning that helps us prevent the early training-time memorization phenomenon. 
Our main contributions are summarized as follows:    
\begin{itemize}
\vspace{-1mm}
\setlength\itemsep{0.2mm}
    \item We introduce a novel SFDA technique that focuses on noise-free self-training exploiting the reliability of generated pseudo-labels. With the help of curriculum learning, we aim to prevent early training time memorization of noisy pseudo-labels and improve the quality of subsequent self-training as shown in Fig.~\ref{fig:main}.
    \item By prioritizing the learning from highly reliable pseudo-labels first, we aim to propagate \emph{refined and accurate label information} among less reliable samples. Such a selective self-training strategy eliminates the requirement of a computationally costly and memory-bank dependent label refinement framework. 
    \item C-SFDA achieves state-of-the-art performance on major benchmarks for image recognition and semantic segmentation. Being highly memory-efficient, the proposed method is also applicable to online test-time adaptation settings and obtains SOTA performance.  
\end{itemize}




\vspace{-1mm}
\section{Related Work}
\vspace{-1mm}
\noindent \textbf{UDA:} 
UDA for visual recognition tasks has been widely studied in the literature \cite{csurka2017comprehensive, wang2018deep}. Adversarial learning~\cite{hoffman2018cycada, tzeng2017adversarial, vu2019advent, long2018conditional}, image-to-image translation~\cite{murez2018image, lee2018diverse, hoffman2018cycada}, cross-domain divergence minimization~\cite{li2020domain, chen2020homm, sun2016deep, shen2018wasserstein},and optimal transport~\cite{damodaran2018deepjdot, chen2020graph, RWOT} are popular techniques across prior works on UDA. Self-training~\cite{xie2020self, feng2021complementary, zou2018unsupervised, mei2020instance, yu2021dast} has recently been a dominant trend in UDA, which uses labeled source data and pseudo-labeled target data (typically generated using a teacher model) to iteratively train a student model. Different variants of self-training such as cycle self-training~\cite{liu2021cycle}, PL Curriculum Learning~\cite{choi2019pseudo}, Selective pseudo-labelling~\cite{wang2020unsupervised} have been proposed to utilize pseudo-labeling in effective manners. Although previous works employed selective pseudo-labeling and curriculum learning in UDA, we aim to exploit these two mechanisms under a new and more challenging scenario (SFDA). This requires us to devise a unique framework for curriculum learning, the effectiveness of which depends on our proposed selective pseudo-labeling technique.   

\noindent \textbf{SFDA:} In recent years, several approaches~\cite{SHOT,3CGAN,liu2021source,A2Net,G-SFDA,universalSFDA,VDM-DA,xu2021learning, prabhu2022augmentation, chen2022contrastive, ding2022source} addressing the source-free domain adaptation problem has been proposed. SHOT~\cite{SHOT} utilizes a centroid-based label refinement technique that guides the self-training. G-SFDA~\cite{G-SFDA} and NRC~\cite{yang2021exploiting} follow a similar strategy with further measures for refining pseudo-labels by encouraging consistent predictions between local neighbor samples. In addition to label refinement, AdaContrast~\cite{chen2022contrastive} leverages MoCo~\cite{he2020momentum} like contrastive feature learning for SFDA by excluding the same class negative pairs detected by the pseudo-labels. However, pseudo-labels generated at the early training stage can be noisy, a fact that has not been well-addressed in these works. Moreover, almost all of these label refinement strategies require having a large memory queue which is undesirable in many real-world scenarios. 
In contrast, our proposed method reduces the risk of label noise memorization without any rigorous and expensive pseudo-label refinement technique.

\begin{figure}[t]
    \centering
    \includegraphics[width=0.95\linewidth]{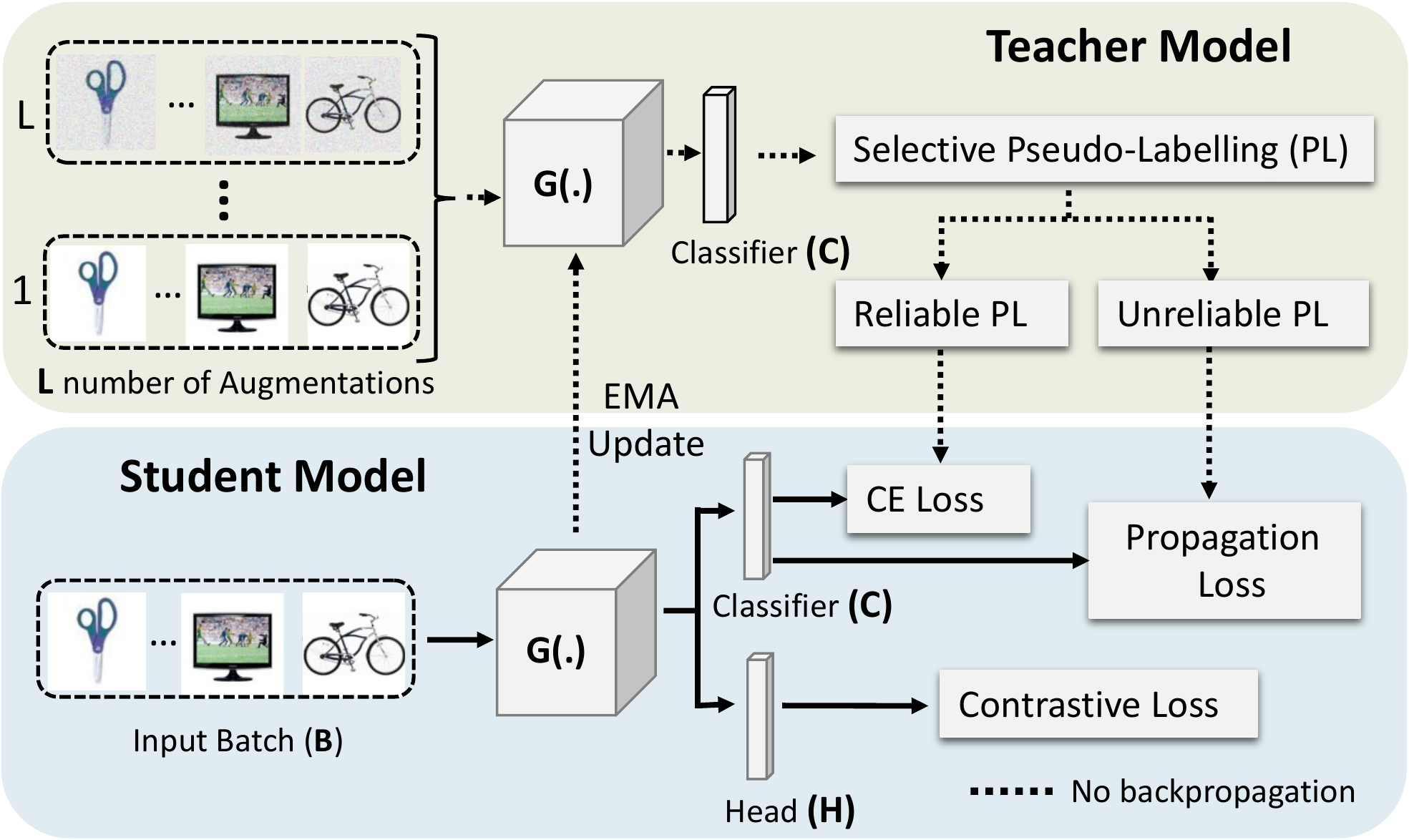}
    \caption{\footnotesize \textbf{Overview of our proposed method} where we use teacher predictions to train the student model. At the early stage of training, we only consider reliable pseudo-labels for CE loss as well as contrastive loss for unsupervised feature representation learning. As the model gets more confident, we use label propagation loss to distribute high-quality label information, learned from reliable pseudo-labels, among unreliable samples. }
    \label{fig:framework}
    \vspace{-3mm}
\end{figure}
\section{Proposed Method}
We start by defining the source domain dataset,   $\mathcal{D}_\mathsf{s}=\{(x_\mathsf{s}^i,y_\mathsf{s}^i )\}_{i=1}^{N_\mathsf{s}}$ containing $N_s$ labeled samples distributed over $K$ number of classes. Here, $y\in\mathcal Y_\mathsf{s}\subseteq\mathbb R^{K} $ denotes the one-hot ground-truth label for sample $x$. Let $f_{\theta_\mathsf{s}}: x_s \rightarrow y_s$ be a source model trained on $\mathcal{D}_\mathsf{s}$. In general, model $f$ consists of a feature extractor $\mathbf{G}$ and a fully connected (FC)-layer based classifier $\mathbf{C}$.  
We denote the target domain dataset, $\mathcal{D}_\mathsf{t}=\{(x_\mathsf{t}^i)\}_{i=1}^{N_\mathsf{t}}$ containing $N_\mathsf{t}$ unlabeled samples. We have access to the target labels $\{y_\mathsf{t}^i)\}_{i=1}^{N_\mathsf{t}}$ for evaluation only. $\mathcal{D}_\mathsf{s}$ and $\mathcal{D}_\mathsf{t}$ have same underlying label distribution with a common label set  $\mathcal{C}=\{1,2,\cdots,K\}$. In this paper, we focus on SFDA problem where access to $\mathcal{D}_\mathsf{s}$ is unavailable while adapting $f_{\theta_\mathsf{s}}$ on $\mathcal{D}_\mathsf{t}$. 
\vspace{-0.5mm}
\subsection{Curriculum SFDA}
We employ a pseudo-labeling based self-training mechanism for SFDA. Consider a self-training framework shown in Fig.~\ref{fig:framework} where we use a teacher model ($f_{\Hat{\theta}_\mathsf{t}}$) for generating the target pseudo-label for a student model ($f_{\theta_\mathsf{t}}$). At the start of training, both models share the same weights, \ie $\theta_\mathsf{t} = \Hat{\theta}_\mathsf{t} = \theta_\mathsf{s}$. For generating the target pseudo-label, we simply use the teacher model prediction, $\Hat{y}_\mathsf{t}$. 
Considering a batch size of $B$, we use the batch of pseudo-labels $\mathcal{Y}_\mathsf{t} = \{\Hat{y}_\mathsf{t}^{(i)}\}_{i=1}^{B}$ to update $\theta_\mathsf{t}$  using a cross-entropy (CE) loss,
\begin{equation}
    \vspace{-0.01cm}
    \mathcal{L}_{ce} = - \frac{1}{B}\sum_{i=1}^{B}{\Hat{y}_\mathsf{tc}^i \cdot \log f_{\theta_\mathsf{t}}(x_\mathsf{t}^i))}, 
    \vspace{-1mm}
\end{equation}
where $\Hat{y}_\mathsf{tc}^i$ is one hot encoding of $\Hat{y}_\mathsf{t}^i$. Minimizing $\mathcal{L}_{ce}$ enforces consistency between student and teacher predictions. 

In our self-training scenario, there is an inevitable presence of noisy labels in $\mathcal{Y}_\mathsf{t} \in \mathbb{R}^{B}$ which often leads to non-optimal model performance \cite{arpit2017closer}. As an initial label-noise preventive measure, we intend to produce stable and high-quality pseudo-labels following  these two simple steps: \emph{i) augmentations averaged prediction, and ii) weight averaged teacher model}. We execute the first step by taking the augmentation-averaged teacher predictions,  
\begin{equation}
    \Hat{y}_\mathsf{t} = \argmax  \frac{1}{L} \sum_{l=1}^{l=L} \Hat{h}^l_t = \argmax \frac{1}{L} \sum_{l=1}^{l=L} f_{\Hat{\theta}_\mathsf{t}}(\hat{x}_\mathsf{t}^l).
\end{equation}
Here, $\hat{x}_\mathsf{t}^l$ is the $l^{\mathsf{th}}$ augmented copy of the input $x_\mathsf{t}$ and $L$ \emph{(=12 in our case)} is the total number of augmentations. Data augmentation~\cite{cubuk2019autoaugment} has been widely used to improve model generalization performance to unseen data \cite{shorten2019survey,krizhevsky2017imagenet}. In our case, however, we aim to generate consistent teacher model predictions over augmented input distributions.
Although augmentations can be dataset-specific and manually designed~\cite{shorten2019survey,krizhevsky2017imagenet, SHOT}, we use a general augmentation policy that is applicable to multiple datasets. For the second step, we simply take the exponential moving average (EMA) of student weights to update the teacher model during each iteration ($j \rightarrow j+1$), 
\begin{equation}
    \Hat{\theta}_{t}^{j+1} =  \gamma \Hat{\theta}_{t}^{j} + (1-\gamma) \theta_{t}^{j+1}, \label{ema}
\end{equation}
where $\gamma$ is a smoothing factor that controls the degree of change we want during each iteration. However, regardless of these measures, adopting an entire batch of pseudo-labels for self-training eventually lead to the memorization of noisy labels. To alleviate this, we propose a curriculum learning aided self-training strategy that encourages learning from high reliable pseudo-labels. 

\vspace{-2mm}
\subsubsection{Selective Pseudo-Labelling}
\label{sec:selection}
\vspace{-1mm}
We describe the process of reliable label selection first as it is a vital part of the proposed curriculum learning strategy. To measure the label reliability, we intend to exploit two widely-used~\cite{dong2021confident,qiao2021uncertainty,wang2021uncertainty,hubalanced} statistics: \emph{i) prediction confidence and ii) average entropy or uncertainty}. In general, if a model is well-calibrated, the accuracy of that model can be strongly related to prediction confidence~\cite{guillory2021predicting}. Therefore, prediction confidence can be a reliable measure of pseudo-label accuracy in SFDA settings. In addition to the confidence score, \cite{guillory2021predicting} shows that difference in entropy can be a reliable estimate for different types of domain shift. While \cite{guillory2021predicting} assumes the availability of both source and target domains, we are restricted to target domain data only. Therefore, we use a carefully designed augmentation policy to create a virtual distribution shift among target domain data. Prediction variance or uncertainty over augmented distributions should give us a close estimate of actual domain shift. To this end, we assign a binary reliability score (${r}^i$) to each target sample based on their prediction confidence and prediction uncertainty ($g_u^i$),
\begin{equation}
    \vspace{-2mm}
    {r}^i = \begin{cases}
      1, & \text{if} \hspace{1mm} \textit{  conf}(\Hat{h}_\mathsf{t}^i) \geq \tau_c \hspace{2mm} \text{and} \hspace{2mm} g_u^i \leq \tau_u\\ 
      0, & \text{otherwise}.\label{eq:reliability}
    \end{cases} 
    \vspace{-1mm}
\end{equation}
We calculate $g_u^i$ by taking the standard deviation (\textit{std.}) over augmentation-based predictions, $g_u^i = \textit{std}\{conf(\Hat{h}_\mathsf{t}^l)\}_{l=1}^{l=L}$. We particularly consider aleatoric uncertainty~\cite{hullermeier2021aleatoric} here since it better addresses the concern of domain shift. The pair of selection thresholds $\tau_c$ and $\tau_u$ can be estimated as 
\begin{equation}
    \vspace{-1mm}
    \tau_c = \frac{1}{B}\sum_{i=1}^{i=B} \textit{  conf}(\Hat{h}_\mathsf{t}^i); \hspace{3mm} \tau_u = \frac{1}{B}\sum_{i=1}^{i=B} g_u^i.\label{eq:con_thr}
    \vspace{-1mm}
\end{equation}
Taking the average as a threshold eliminates the requirement of per-dataset hyper-parameter tuning and makes our selection process highly adaptive. Note that, the proposed selection strategy is also applicable to fully test-time adaptation~\cite{wang2021tent,chen2022contrastive}. 
\vspace{-2mm}
\subsubsection{Loss Functions}
\vspace{-1mm}
After getting the reliability score for each sample, we separate the input batch $\mathbb{D}$ into \emph{more reliable (R)} and \emph{less reliable (U)} groups, $\mathbb{D}_R=\{(x_\mathsf{t}^i,\Hat{y}_\mathsf{t}^i):r_i=1\}_{i=1}^{B}$ and $\mathbb{D}_U=\{(x_\mathsf{t}^i,\Hat{y}_\mathsf{t}^i):r_i=0\}_{i=1}^{B}$. While this gives us a good estimate of reliable samples, $\mathbb{D}_R$ may lack diverse samples (sometimes, missing some categories completely). As a potential remedy to this, we choose a few samples from $\mathbb{D}_U$ based on another metric: \emph{Top-2 confidence score difference (DoC)} and consider them as reliable. Finally, we employ class-balanced cross-entropy loss for $\mathbb{D}_R$ ($\mathcal{L}_{ce}^R$) with an inverse frequency loss-weighting factor ($\lambda_k$) that accounts for the label imbalance in $\mathbb{D}_R$. Details of DoC and $\lambda_k$ are in supplementary. For $\mathbb{D}_U$, we employ label propagation loss\cite{zhou2003learning} as follows,
\begin{equation}
   \vspace{-1mm}
    \mathcal{L}_\mathcal{P} = \frac{1}{2|\mathbb{D}_U|} \sum_{i=1}^{|\mathbb{D}_U|} \Vert f_{\theta_\mathsf{t}}(x_\mathsf{t}^i) - \Hat{y}_\mathsf{tc}^i \Vert^2 .\label{eq:label_prop} 
    \vspace{-0.5mm}
\end{equation}
Due to the transductive property of $\mathcal{L}_\mathcal{P}$, it propagates label information from $\mathbb{D}_R$ to $\mathbb{D}_U$. 

Note that both $\mathcal{L}_{CE}^{R}$ and 
$\mathcal{L}_{P}$ require pseudo-label which may lead to memorization; depending on the success in selection stage. In addition to supervised self-training, learning useful representations of images in an unsupervised manner may reduce the risk of memorization. One such approach is \emph{fully unsupervised contrastive learning (CL)} where meaningful representation learning becomes possible by enforcing similarity between two augmented copies of each sample $x_t$, ${x_t}^{aug,1}$ and ${x_t}^{aug,2}$. To this end, we employ a projection head $\mathbf{H}$ to obtain feature projections $q_{i} = \mathbf{H}(\mathbf{G}({x_t}^{aug,1}))$, and $q_{j} = \mathbf{H}(\mathbf{G}({x_t}^{aug,2}))$ that gives us the contrastive criterion \cite{chen2020simple,khosla2020supervised} as
\begin{align}
        \vspace{-1mm}
        \ell_{i,j}  & = -\log \frac{\exp(\mathrm{sim}(q_{i}, q_{j})/\kappa)}{\sum_{b=1}^{2B} 1_{b \neq i}\exp(\mathrm{sim}(q_i, q_b)/\kappa)}~, \\
        & \mathcal{L}_\mathcal{C}= \frac{1}{2B}\sum_{b=1}^{2B}[\ell_{2b-1,2b} + \ell_{2b,2b-1}] \label{eq:contratstive_loss},
        \vspace{-1mm}
\end{align}
where $1_{b \neq i}$ is an indicator function that gives a 1 if $b \neq i$, $\kappa$ is a temperature constant and $\mathrm{sim}(q_{i}, q_{j})$ is the cosine similarity between $q_{i}$ and $q_{j}$. Even though label-dependent contrastive learning has been employed for SFDA\cite{chen2022contrastive}, we focus on label-independent CL to minimize the effect of label noise; especially at an early stage of training. Finally, the total loss can be expressed as 
\begin{equation}
    \mathcal{L}_{tot} =  \mu_r \mathcal{L}_{ce}^R + (1-\mu_r) \mathcal{L}_\mathcal{P} + \mu_c \mathcal{L}_\mathcal{C}, \label{eq:total_loss} 
\end{equation}
where $\mu_r$ and $\mu_c$ are loss coefficients that dictate the pace of curriculum learning we propose next. 

\vspace{-2mm}
\subsubsection{Curriculum Learning}
\vspace{-1mm}
Curriculum Learning \cite{bengio2009curriculum,zhou2018minimax} promotes the strategy of learning from easier samples first and harder samples later. Our selection strategy in Section~\ref{sec:selection} provides us with an estimation of easy and hard groups. Since pseudo-labels in $\mathbb{D}_R$ are most likely to be correct, DNN finds it easier to learn from them. On the other hand, learning from $\mathbb{D}_U$ should be more restricted due to the presence of a higher noise level. Therefore, we set an update equation for $\mu_r$ as 
\begin{equation}
    \mu_r^{j} = \mu_r^{j-1}(1- \alpha e^{-\frac{1}{d^{j}}}),\label{eq:loss_coeff}
\end{equation}
where $d^{j} = \frac{\tau_u}{\tau_c}$ is difficulty score of current batch of samples and $\mu_r^{j-1}$ is the labeled loss coefficient at previous iteration. We set $\alpha$ and $\mu_r^{0}$ to 0.005 and 1, respectively, to restrict the learning from $\mathbb{D}_U$ since pseudo-labels are mostly noisy during the early stages of training. As the training progresses and overall reliability improves, we start learning from $\mathbb{D}_U$ by gradually decreasing $\mu_r$. In addition, the change in $\mu_r$ is directly controlled by the difficulty in learning the current batch of samples. If the batch of samples at iteration $j$ is hard-to-learn, (\ie $d^j$ is high), we keep the change in $\mu_r$ to minimal. Similarly, we exponentially decrease the contrastive loss coefficient $\mu_c$ as  
\begin{equation}
    \mu_c^{j} = \mu_c^{j-1}e^{-\beta}.\label{eq:con_loss_coeff}
\end{equation}
We set initial $\mu_c^{0}$ to 0.5 as unsupervised feature learning helps more at early stages of training. $\beta$ is set to be 1e-4. 

\vspace{-2mm}
\subsubsection{Semantic Segmentation}
\vspace{-1mm}
Up to now, we have only considered the classification task where each input sample is associated with a single label. However, semantic segmentation is a multi-label classification task where we assign a label to each pixel. Consider a target domain image $x \in \mathbb{R}^{H \times W}$ where $x_{ij}$ indicates the pixel of $i^{th}$ row and $j^{th}$ column. The task at hand is to assign one of $K$ semantic labels, $y_{ij} \in {1,2, \ldots K}$ to each $x_{ij}$. For $x$, we use a model ($f$) to produce a probabilistic output prediction $p \in \mathbb{R}^{H \times W \times K}$ over $K$ classes. The map of pseudo-labels ($\hat{y} \in \mathbb{R}^{H \times W}$)can be estimated as $\hat{y} = \argmax_{k} p; k \in {1,2, \ldots K}$. As some predictions are more reliable than others, using similar selection criteria (as image classification) to separate pixels makes sense. However, instead of using one single threshold for all pixels, we instead choose per-category thresholds. To this end, we estimate a pair of thresholds for each category $k$. Given a batch, we accumulate all confidence scores and select the per-category confidence threshold, $\tau_c^k$, as the P-th percentile confidence score. Similarly, we select P-th percentile uncertainty score for the uncertainty threshold, $\tau_c^k$. In our work, we set the value of P to 55. After choosing the thresholds, we follow eq.~\ref{eq:reliability} to assign a per-pixel reliability score, $r_{ij}$. As for loss functions, we consider cross-entropy loss ($\mathcal{L}^R_{ce}$) for the reliable labels $\hat{y}^R$, and to promote diverse predictions, we minimize the prediction entropy loss,
\begin{equation}
    \vspace{-0.5mm}
    \mathcal{L}_{E} = - \frac{1}{HW}\sum_{i=1,j=1}^{H,W} p_{ij} \cdot \log(p_{ij}). \label{eq:entropy}
    \vspace{-0.5mm}
\end{equation}
Finally, we update our model by minimizing the total loss,
\begin{equation}
    \mathcal{L}_{tot} = \mathcal{L}^R_{ce} + \mu_{e} \mathcal{L}_{E},
\end{equation}
Where $\mu_{e}$ is the entropy loss coefficient. We follow a similar update equation as \ref{eq:con_loss_coeff} for $\mu_E$ with an initial value of, $\mu^0_e$ = 1e-3. Note that, we only update BN layers and freeze other parameters. For uncertainty measures, we use ColorJitter and Gaussian noise as augmentation transformations. More details are in the supplementary.

\section{Experiments}
\label{sec:experiments}
\subsection{Experimental Settings}
\noindent \textbf{Image Classification Datasets:} \emph{Office-31}~\cite{office31} is a small-scale benchmark with images from 31 categories across 3 domains, \textbf{A}mazon (2,817), \textbf{D}SLR (498) and \textbf{W}ebcam (795). \emph{Office-Home}~\cite{venkateswara2017deep} has a total of 15.5K images from 65 classes  collected from 4 different image domains: \textbf{Ar}tistic, \textbf{Cl}ipart, \textbf{Pr}oduct, and \textbf{R}eal-\textbf{w}orld. We consider 12 transfer tasks for this dataset. \emph{VisDA}~\cite{peng2017visda} contains 2 different domains, synthetic and real, with 12 classes in both domains. The synthetic or source domain contains 150K rendered 3D images with different poses. The corresponding real or target domain contains about 55K real-world images. For evaluation, we consider per-class accuracy and the average (Avg.) over them.  \emph{DomainNet}~\cite{peng2019moment} is another large-scale dataset with 6 domains containing over 500K images from 126 classes. We consider 4 domains (\textbf{R}eal, \textbf{S}ketch, \textbf{C}lipart, \textbf{P}ainting), as \cite{saito2019semi} identify severe noisy labels in the dataset. We evaluate the methods on 7 transfer tasks between 4 domains and report top-1 accuracy. 
\begin{table}[tb]
\caption{\footnotesize Classification accuracy (\%) under UDA and SFDA settings on \textbf{Office-31} dataset (ResNet50 backbone). We report Top-1 accuracy on 6 domain shifts ($\to$) and take the average (Avg.) over them. The best results under the SFDA setting are shown in bold font.}
\vspace{-2mm}
\newcommand{\tabincell}[2]{\begin{tabular}{@{}#1@{}}#2\end{tabular}}
\centering
\resizebox{\linewidth}{!}{
\begin{tabular}{c|c|cccccc|c}
\toprule
Method & \tabincell{c}{source\\-free} & 
A$\rightarrow$D & A$\rightarrow$W & D$\rightarrow$A & D$\rightarrow$W & W$\rightarrow$A & W$\rightarrow$D & \cellcolor{lightgray!30}Avg. \\
\midrule
GSDA\cite{GSDA} & $\times$ & 94.8 & 95.7 & 73.5 & 99.1 & 74.9 & 100 & \cellcolor{lightgray!30}89.7\\
CAN\cite{CAN} & $\times$ & 95.0 & 94.5 & 78.0 & 99.1 & 77.0 & 99.8 & \cellcolor{lightgray!30}90.6\\
SRDC\cite{SRDC} & $\times$ & 95.8 &  95.7 & 76.7 & 99.2 & 77.1 & 100 & \cellcolor{lightgray!30}90.8\\
\midrule
SHOT\cite{SHOT}  & \checkmark & 94.0 & 90.1 & 74.7 & 98.4 & 74.3 & 99.9 & \cellcolor{lightgray!30}88.6\\
3C-GAN\cite{3CGAN}& \checkmark & 92.7 & 93.7 & 75.3 & 98.5 & 77.8 & 99.8 & \cellcolor{lightgray!30}89.6\\
A$^2$Net\cite{A2Net} & \checkmark & 94.5 & 94.0 & 76.7 & \textbf{99.2} & 76.1 & \textbf{100} & \cellcolor{lightgray!30}90.1\\
SFDA-DE~\cite{ding2022source}  & \checkmark & 96.0 & \textbf{94.2} & 76.6 & 98.5 & 75.5 & 99.8 & \cellcolor{lightgray!30}90.1\\
\rowcolor{aliceblue} C-SFDA (Ours)  & \checkmark & \textbf{96.2} & 93.9 & \textbf{77.3} & 98.8 & \textbf{77.9} & 99.7 & \cellcolor{lightgray!30}\textbf{90.5}\\
\bottomrule
\end{tabular}
}
\label{office31result}
\vspace{-2.5mm}
\end{table}

\vspace{0.7mm}

\noindent \textbf{Semantic Segmentation Datasets:}
For segmentation, we consider GTA5$\to$Cityscapes, SYNTHIA$\to$Cityscapes \& CityScapes$\to$Dark-Zurich adaptations tasks. GTA5~\cite{richter2016playing} contains $\sim$25k synthetic images, with a resolution of $1914\times1052$, generated from GTA5 video frames. Cityscapes~\cite{cordts2016cityscapes} provides 3,975 daytime street scenes, with a resolution of $2048\times1024$, from 50 different cities. Following prior work~\cite{hoffman2018cycada,vu2019advent,zou2019confidence}, we consider splitting Cityscapes images into train-val splits and report 19-way classification performance over the validation split. SYNTHIA~\cite{ros2016synthia} is another synthetic dataset with 9400 scenes of size 1280x760. As SYNTHIA and Cityscapes have overlaps only for 16 categories, we report 16-way and 13-way performances for SYNTHIA$\to$Cityscapes. Dark Zurich~\cite{sakaridis2019guided} is a large dataset with 2,416 nighttime unlabeled images of 1080p resolution.


\renewcommand{\arraystretch}{0.9}
\begin{table*}[htb]
\caption{\footnotesize Classification performance (\%) under UDA and SFDA settings on \textbf{Office-Home} dataset (ResNet50 backbone). We report Top-1 accuracy on 12 domain shifts ($\to$) and take the average (Avg.) over them. Our method achieves SOTA performance on 8 of these shifts.}
\vspace{-2mm}
\centering
\resizebox{0.87\linewidth}{!}{
\begin{tabular}{c|c|cccccccccccc|c}
\toprule
Method & SF & 
Ar$\rightarrow$Cl & Ar$\rightarrow$Pr & Ar$\rightarrow$Rw & Cl$\rightarrow$Ar & Cl$\rightarrow$Pr & Cl$\rightarrow$Rw & Pr$\rightarrow$Ar & Pr$\rightarrow$Cl & Pr$\rightarrow$Rw & Rw$\rightarrow$Ar & Rw$\rightarrow$Cl & Rw$\rightarrow$Pr & \cellcolor{lightgray!30}Avg. \\
\midrule
RSDA\cite{RSDA} & $\times$ & 53.2 & 77.7 & 81.3 & 66.4 & 74.0 & 76.5 & 67.9 & 53.0 & 82.0 & 75.8 & 57.8 & 85.4 & \cellcolor{lightgray!30}70.9 \\
TSA\cite{li2021transferable} & $\times$ & 57.6 & 75.8 & 80.7 & 64.3 & 76.3 & 75.1 & 66.7 & 55.7 & 81.2 & 75.7 & 61.9 & 83.8 & \cellcolor{lightgray!30}71.2\\
SRDC\cite{SRDC} & $\times$ & 52.3 & 76.3 & 81.0 & 69.5 & 76.2 & 78.0 & 68.7 & 53.8 & 81.7 & 76.3 & 57.1 & 85.0 & \cellcolor{lightgray!30}71.3 \\
FixBi\cite{FixBi} & $\times$ & 58.1 & 77.3 & 80.4 & 67.7 & 79.5 & 78.1 & 65.8 & 57.9 & 81.7 & 76.4 & 62.9 & 86.7 & \cellcolor{lightgray!30}72.7 \\
\midrule
G-SFDA\cite{G-SFDA} & \checkmark & 57.9 & 78.6 & 81.0 & 66.7 & 77.2 & 77.2 & 65.6 & 56.0 & 82.2 & 72.0 & 57.8 & 83.4 & \cellcolor{lightgray!30}71.3\\
SHOT\cite{SHOT} & \checkmark & 57.1 & 78.1 & 81.5 & 68.0 & 78.2 & 78.1 & 67.4 & 54.9 & 82.2 & 73.3 & 58.8 & 84.3 & \cellcolor{lightgray!30}71.8\\
HCL~\cite{huang2021model} & \checkmark & \textbf{64.0} & 78.6 & 82.4 & 64.5 & 73.1 & 80.1 & 64.8 & \textbf{59.8} & 75.3 & \textbf{78.1} & \textbf{69.3} & 81.5 & \cellcolor{lightgray!30}72.6 \\
A$^2$Net\cite{A2Net} & \checkmark & 58.4 & 79.0 & 82.4 & 67.5 & 79.3 & 78.9 & \textbf{68.0} & 56.2 & 82.9 & 74.1 & 60.5 & 85.0 & \cellcolor{lightgray!30}72.8\\
SFDA-DE~\cite{ding2022source} & \checkmark & 59.7 & 79.5 & 82.4 & \textbf{69.7} & 78.6 & \textbf{79.2} & 66.1 & 57.2 & 82.6 & 73.9 & 60.8 & 85.5 & \cellcolor{lightgray!30}72.9\\
\rowcolor{aliceblue} C-SFDA (Ours) & \checkmark & 60.3 & \textbf{80.2} & \textbf{82.9} & 69.3 & \textbf{80.1} & 78.8 & 67.3 & 58.1 & \textbf{83.4} & 73.6 & 61.3 & \textbf{86.3} & \cellcolor{lightgray!30}\textbf{73.5}\\
\bottomrule
\end{tabular}}
\label{officehomeresult}
\vspace{-1mm}
\end{table*}
\renewcommand{\arraystretch}{1}

\renewcommand{\arraystretch}{0.9}
\begin{table*}[htb]
\caption{\footnotesize Source-free (SF) domain adaptation performance on \textbf{VisDA} dataset  (ResNet-101 backbone) shown by per-class accuracy (\%) and their average (Avg.). Our method improves the average accuracy by 1\% compared to the previous SOTA, Adacon~\cite{chen2022contrastive}. }
\vspace{-2mm}
\footnotesize
\centering
\scalebox{0.85}{
\begin{tabular}{c|c|cccccccccccc|c}
\toprule
Method & SF & plane &  bike &  bus &  car &  horse &  knife & mcycle &  person &  plant &  sktbrd &  train &  truck &  \cellcolor{lightgray!30}Avg. \\
\midrule
MCC\cite{MCC} & $\times$ & 88.7 & 80.3 & 80.5 & 71.5 & 90.1 & 93.2 & 85.0 & 71.6 & 89.4 & 73.8 & 85.0 & 36.9 & \cellcolor{lightgray!30}78.8\\
STAR\cite{STAR} & $\times$ & 95.0 & 84.0 & 84.6 & 73.0 & 91.6 & 91.8 & 85.9 & 78.4 & 94.4 & 84.7 & 87.0 & 42.2 & \cellcolor{lightgray!30}82.7 \\
RWOT\cite{RWOT} & $\times$ & 95.1 & 80.3 & 83.7 & 90.0 & 92.4 & 68.0 & 92.5 & 82.2 & 87.9 & 78.4 & 90.4 & 68.2 & \cellcolor{lightgray!30}84.0 \\
SE\cite{SE} & $\times$ & 95.9 & 87.4 & 85.2 & 58.6 & 96.2 & 95.7 & 90.6 & 80.0 & 94.8 & 90.8 & 88.4 & 47.9 & \cellcolor{lightgray!30}84.3\\

\midrule
Source only & - & 57.2 & 11.1 & 42.4 & 66.9 & 55.0 & 4.4 & 81.1 & 27.3 & 57.9 & 29.4 & 86.7 & 5.8 & \cellcolor{lightgray!30}43.8 \\
SHOT\cite{SHOT} & \checkmark & 94.3 & 88.5 & 80.1 & 57.3 & 93.1 & 94.9 & 80.7 & 80.3 & 91.5 & 89.1 & 86.3 & 58.2 & \cellcolor{lightgray!30}82.9 \\
A$^2$Net\cite{A2Net} & \checkmark & 94.0 & 87.8 & 85.6 & 66.8 & 93.7 & 95.1 & 85.8 & 81.2 & 91.6 & 88.2 & 86.5 & 56.0 & \cellcolor{lightgray!30}84.3 \\
SFDA-DE~\cite{ding2022source} & \checkmark & 95.3 & \textbf{91.2} & 77.5 & 72.1 & 95.7 & \textbf{97.8} & 85.5 & \textbf{86.1} & \textbf{95.5} & \textbf{93.0} & 86.3 & 61.6 & \cellcolor{lightgray!30}86.5 \\
AdaCon~\cite{chen2022contrastive} & \checkmark  & 97.0 & 84.7 & 84.0 & \textbf{77.3} & 96.7 & 93.8 & 91.9 & 84.8 & 94.3 & 93.1 & \textbf{94.1} & 49.7 & \cellcolor{lightgray!30}86.8 \\

\rowcolor{aliceblue} C-SFDA (Ours) & \checkmark & \textbf{97.6} & 88.8 & \textbf{86.1} & 72.2 & \textbf{97.2} & 94.4 & \textbf{92.1} & 84.7 & 93.0 & 90.7 & 93.1 & \textbf{63.5} & \cellcolor{lightgray!30}\textbf{87.8} \\
\midrule
AdaCon~\cite{chen2022contrastive} (Online) & \checkmark & 95.0 & 68.0 & 82.7 & \textbf{69.6} & 94.3 & 80.8 & 90.3 & 79.6 & \textbf{90.6} & 69.7 & \textbf{87.6} & 36.0 & \cellcolor{lightgray!30}78.7 \\
\rowcolor{aliceblue} C-SFDA (Online) & \checkmark & \textbf{95.9} & \textbf{75.6} & \textbf{88.4} & 68.1 & \textbf{95.4} & \textbf{86.1} & \textbf{94.5} & \textbf{82.0} & 89.2 & \textbf{80.2} & 87.3 & \textbf{43.8} & \cellcolor{lightgray!30}\textbf{82.1} \\
\bottomrule
\end{tabular}}
\label{visdaresult}
\vspace{-2mm}
\end{table*}

\vspace{0.7mm}

\noindent \textbf{Implementation Details:} We use ResNet50~\cite{he2016deep} backbone for Office-31, Office-Home, DomainNet and ResNet-101~\cite{he2016deep} for VisDA. Following SHOT~\cite{SHOT}, we replace the fully connected (FC) layer with a 256-dimensional bottleneck layer and task-specific FC classification layer. We use batch normalization~\cite{ioffe2015batch} after bottleneck and apply WeightNorm~\cite{salimans2016weight} on the classifier. For source training, we initialize the models with ImageNet-1K~\cite{deng2009imagenet} pre-trained weights. Following \cite{SHOT}, we split the source dataset into the train (90\%) and validation (10\%) sets. We employ a 10 times higher learning rate for bottleneck and classifier than the backbone. For target domain adaptation, we use similar training settings for all the datasets. For Office datasets~\cite{office31,venkateswara2017deep}, we use SGD optimizer with a learning rate of 5e-3, a momentum of 0.9 with a weight decay of 1e-4 and a batch size of 128. For VisDA and DomainNet, we use a learning rate of 5e-4 with cosine annealing~\cite{chen2022contrastive}. We train for 20 epochs with a batch size of 128 for VisDA. We consider a larger batch size of 256 for DomainNet and train for 25 epochs. We set $\eta$ to 0.98 for EMA updates across datasets. 

For all semantic segmentation, we use DeepLabV2~\cite{chen2017deeplab} with a ResNet101~\cite{he2016deep} backbone and initialize models with ImageNet-1K~\cite{deng2009imagenet} pre-trained weights. For GTA5 and SYNTHIA source training, we use an SGD optimizer with a 1e-4 learning rate, a 0.9 momentum, and a weight decay of 5e-4. We train the model for 20 epochs with a batch size of 8 and apply different weather augmentations~\cite{michaelis2019benchmarking} during training. For Cityscapes, we follow the settings in ~\cite{sakaridis2019guided} and use a learning rate of 2.5e-4. During adaptation, we use a learning rate of 1e-4 to tune only the batch normalization (BN) parameters. With a batch size of 8, we train the model for 50K steps. Note that we only consider online adaptation for Cityscapes$\to$Dark-Zurich and train the model for 1 epoch.  Similar to Image classification, we also consider EMA update for segmentation and set $\eta$ to $0.995$. Please see supplementary for more details.


\vspace{-1mm}
\subsection{Experimental Results}
\vspace{-0.5mm}
\noindent \textbf{Evaluation on Image Classification Task:}
We compare the proposed method on Image Classification benchmarks in Table~\ref{office31result}-\ref{tab:2_domainnet_main}. We report the Top-1 accuracy for each domain shift and take their average. For Office-31 dataset, we achieve an average 0.4\% accuracy improvement over the previous SOTA. We also achieve a similar improvement (0.6\%) for the Office-Home dataset. We believe, avoiding the early training time label noise propagation, helps our method significantly to perform well. In VisDA, C-SFDA outperforms SOTA AdaCon~\cite{chen2022contrastive} by 1\% and obtains significant performance improvement for the rare classes such as "truck". Table~\ref{tab:2_domainnet_main} shows that the proposed method sees similar accuracy improvement (1.2\%) over the previous SOTA for DomainNet. Although AdaCon~\cite{chen2022contrastive} uses a large memory queue to refine the pseudo-labels, it still suffers from early training time memorization. Whereas utilizing a \emph{label-selection technique} for curriculum training, C-SFDA eliminates the requirement of a \emph{label-refinement technique} and still outperforms AdaCon~\cite{chen2022contrastive}. We also consider several general UDA techniques considering continued source data access. We encouragingly find that the proposed C-SFDA performs better than most of these methods across datasets, even without source data access. 
\renewcommand{\arraystretch}{1}

\begin{table}[ht]
\centering
\caption{\footnotesize Classification accuracy (\%) on \textbf{DomainNet} for source-free domain adaptation (ResNet-50 backbone). }
\vspace{-2mm}
\resizebox{\columnwidth}{!}{
\begin{tabular}{c|c|ccccccc|cc}
    \Xhline{1pt}
    Method & SF  & R$\to$C & R$\to$P & P$\to$C & C$\to$S & S$\to$P & R$\to$S & P$\to$R & \cellcolor{lightgray!30}Avg. \\
    \hline
    MCC~\cite{MCC} & $\times$ & 44.8  &  65.7   & 41.9 & 34.9 & 47.3 & 35.3 & 72.4 & \cellcolor{lightgray!30}48.9 \\
    \midrule
    Source only & - & 55.5 & 62.7 & 53.0 & 46.9 & 50.1 & 46.3 & 75.0 & \cellcolor{lightgray!30}55.6\\
    TENT~\cite{wang2021tent} & \checkmark & 58.5 & 65.7 & 57.9 & 48.5 & 52.4 & 54.0 & 67.0 & \cellcolor{lightgray!30}57.7 \\
    SHOT \cite{SHOT} & \checkmark & 67.7 & 68.4 & 66.9 & 60.1 & 66.1 & 59.9 & \textbf{80.8} & \cellcolor{lightgray!30}67.1 \\
    AdaCon~\cite{chen2022contrastive} & \checkmark & 70.2 & 69.8 & \textbf{68.6} & 58.0 & 65.9 & 61.5 & 80.5 & \cellcolor{lightgray!30}67.8 \\
    \rowcolor{aliceblue} Ours & \checkmark & \textbf{70.8} & \textbf{71.1} & 68.5 & \textbf{62.1} & \textbf{67.4} & \textbf{62.7} & 80.4 & \cellcolor{lightgray!30}\textbf{69.0} \\    
    \midrule
    AdaCon~\cite{chen2022contrastive} (online) & \checkmark & 61.1 &	66.9 &	60.8 & 53.4	 & 62.7	 & 54.5 &	\textbf{78.9} &	\cellcolor{lightgray!30}62.6 \\
    \rowcolor{aliceblue} Ours (online) & \checkmark & \textbf{61.6} & \textbf{67.4} &	\textbf{61.3} & \textbf{55.1} & \textbf{63.2} & \textbf{54.8} & 78.5 &	\cellcolor{lightgray!30}\textbf{63.1} \\
    \bottomrule
\end{tabular}
}
\label{tab:2_domainnet_main}
\vspace{-2mm}
\end{table}

\renewcommand{\arraystretch}{0.9}
\begin{table*}[t]
	\centering
	\caption{\footnotesize
	Performance evaluation on \textbf{GTA5$\to$Cityscapes} (DeepLabV2 with ResNet101) where we report mean IoU (mIoU) over 19 categories on Cityscapes validations set. Our method achieves the best mIoU in SFDA and online test-time adaptation.}
	\vspace{-2mm}
	\resizebox{\linewidth}{!}{
	\begin{tabular}{c|c|ccccccccccccccccccc|c}
		\toprule
		Method  &SF & Road & SW & Build & Wall & Fence & Pole & TL & TS & Veg. & Terrain & Sky & PR & Rider & Car & Truck & Bus & Train & Motor & Bike & \cellcolor{green!5}mIoU\\
		\midrule
		CrCDA~\cite{huang2020contextual} &$\times$ &{92.4}	&55.3	&{82.3}	&31.2	&{29.1}	&32.5	&33.2	&{35.6}	&83.5	&34.8	&{84.2}	&58.9	&{32.2}	&84.7	&{40.6}	&46.1	&2.1	&31.1	&32.7	&\cellcolor{green!5}48.6\\
		ProDA~\cite{zhang2021prototypical}   &$\times$  & 91.5 & 52.4    & 82.9     & {42.0} & { 35.7}  & 40.0 & 44.4  & 43.3 & { 87.0}       & { 43.8}    & 79.5 & 66.5   & 31.4  & 86.7 & 41.1  & 52.5 & 0.0   & 45.4      & {53.8} & \cellcolor{green!5}53.7 \\  
		CPSL~\cite{li2022class}  &$\times$  &  91.7     &  52.9   &  83.6   &   {43.0}   &    32.3   &  {43.7}    &   {51.3}    &    42.8        &   85.4      &   37.6   &   81.1     &   { 69.5}    &  30.0    &   {88.1}    &   44.1   & { 59.9}      &     24.9      &  { 47.2}  &  48.4  & \cellcolor{green!5}55.7   \\
		\midrule
		Source Only & - & 69.7&20.5 &73.3 &22.1 & 12.3 & 23.5 & 31.8& 17.9 &78.7 & 18.7 & 68.2 & 53.9 & 26.5  & 70.6 & 32.2 & 4.5 & 8.1 & 26.8 & 31.5 & \cellcolor{green!5}36.4 \\
		UR~\cite{sivaprasad2021uncertainty} &\checkmark& 92.3 &55.2 &81.6 &30.8 &18.8 &37.1 &17.7 &12.1 &84.2 &35.9 &83.8 &57.7 &24.1 &81.7 &27.5 &44.3 &6.9 &24.1 &40.4 &\cellcolor{green!5} 45.1\\
		SFDA~\cite{liu2021source} &\checkmark&91.7	&52.7	&82.2	&28.7	&20.3	&36.5	&30.6	&23.6	&81.7	&35.6	&84.8	&59.5	&22.6	&83.4	&29.6	&32.4	&11.8	&23.8	&39.6	& \cellcolor{green!5}45.8\\
		HCL~\cite{huang2021model} &\checkmark& 92.0	& 55.0	& 80.4	& 33.5	& 24.6	&37.1	&35.1	&28.8	&83.0	&37.6	&82.3	&59.4	&27.6	&83.6	&32.3	&36.6	&14.1	&28.7	&43.0	&\cellcolor{green!5}48.1 \\
        \rowcolor{aliceblue}  C-SFDA (ours) & \checkmark & 90.4 & 42.2 & 83.2 & 34.0 & 29.3 & 34.5 & 36.1 & 38.4 & 84.0 & 43.0 & 75.6 & 60.2 & 28.4 & 85.2 & 33.1 & 46.4 & 3.5 & 28.2 & 44.8 & \cellcolor{green!5}\textbf{48.3} \\
        \midrule
        AUGCO~\cite{prabhu2022augmentation} (Online) & \checkmark & 90.3 & 41.2 & 81.8 & 26.5 & 21.4 & 34.5 & 404. & 33.3 & 83.6 & 34.6 & 79.7 & 61.4 & 19.3 & 84.7 & 30.3 & 39.5 & 7.3 & 27.6 & 34.6 & \cellcolor{green!5}45.9 \\
        \rowcolor{aliceblue} C-SFDA (Online) & \checkmark & 84.7 & 37.8 & 82.4 & 29.7 & 28.0 & 31.8 & 34.8 & 29.3 & 83.7 & 43.8 & 76.9 & 58.8 & 28.4 & 84.9 & 33.5 & 44.1 & 0.5 & 24.5 & 39.1 & \cellcolor{green!5}\textbf{46.3}\\
        \bottomrule
        
	\end{tabular}
	}
	\label{table:gta2city}
	\vspace{-1mm}
\end{table*}
\renewcommand{\arraystretch}{1}

\renewcommand{\arraystretch}{0.9}
\begin{table*}[!h]
	\centering
	\caption{\footnotesize
	Performance evaluation on \textbf{SYNTHIA$\to$Cityscapes}. We report mean IoU (mIoU) over 16 common categories between SYNTHIA and Cityscapes. mIoU\textsuperscript{*} are calculated over 13 categories. Our method achieves SOTA performance in both mIoU and mIoU\textsuperscript{*}.
	}
	\vspace{-2mm}
	\resizebox{\linewidth}{!}{
	\begin{tabular}{c|c|cccccccccccccccc|c|c}
		\toprule
		Method  & SF & Road & SW & Build & Wall\textsuperscript{*} & Fence\textsuperscript{*} & Pole\textsuperscript{*} & TL & TS & Veg. & Sky & PR & Rider & Car & Bus & Motor & Bike & \cellcolor{magenta!5}mIoU & \cellcolor{green!5} mIoU\textsuperscript{*}\\
		\midrule
        CrCDA\cite{huang2020contextual} &$\times$ &{86.2}	&{44.9}	&79.5	&8.3	&{0.7}	&{27.8}	&9.4	&11.8	&78.6	&{86.5}	&57.2	&{26.1}	&{76.8}	&{39.9}	&21.5	&32.1	&\cellcolor{magenta!5}{42.9}	&\cellcolor{green!5}{50.0}\\
		ProDA~\cite{zhang2021prototypical}   &$\times$    & 87.1 &44.0   &83.2    & 26.9 & 0.7   & 42.0 & 45.8 & 34.2 & 86.7      & 81.3	& 68.4   & 22.1  & 87.7 & 50.0 & 31.4      & 38.6   & \cellcolor{magenta!5}51.9 & \cellcolor{green!5}58.5\\
		CPSL~\cite{li2022class}   &$\times$  &   87.3   &    44.4     &    83.8      &   25.0   &   0.4    &   42.9   &   47.5   &   32.4   &     86.5       &    83.3  &    69.6   &   29.1    &   89.4  &   52.1  &    42.6     &  54.1  &  \cellcolor{magenta!5}54.4 & \cellcolor{green!5}61.7 \\ 
        \midrule
        Source Only & - & 45.2 & 19.6 &  72.0 & 6.7 & 0.1& 24.3 & 5.5 & 7.8 & 74.4 & 81.9 & 57.3 & 17.3 & 39.0 & 19.5 & 7.0 & 6.2 & \cellcolor{magenta!5}31.3 & \cellcolor{green!5}36.2 \\
        
        UR~\cite{sivaprasad2021uncertainty} &\checkmark&59.3 &24.6 &77.0 &14.0 &1.8 &31.5 &18.3 &32.0 &83.1 &80.4 &46.3 &17.8 &76.7 &17.0 &18.5 &34.6 &\cellcolor{magenta!5}39.6 &\cellcolor{green!5}45.0\\
		SFDA~\cite{liu2021source} &\checkmark&67.8	&31.9	&77.1	&8.3	&1.1	& 35.9	& 21.2	& 26.7	& 79.8 &79.4	&58.8	&27.3	&80.4	&25.3	& 19.5	& 37.4	& \cellcolor{magenta!5}42.4	& \cellcolor{green!5}48.7 \\
        HCL~\cite{huang2021model} &\checkmark&80.9	&34.9	&76.7	&6.6	&0.2	&36.1	&20.1	&28.2	&79.1	&83.1	&55.6	&25.6	&78.8	&32.7	&24.1	&32.7	&\cellcolor{magenta!5}43.5	&\cellcolor{green!5}50.2\\
		\rowcolor{aliceblue} C-SFDA (Ours) & \checkmark& 87.0	& 39.0	& 79.5	& 12.2	& 1.8 & 32.2 & 20.4	& 24.3	& 79.5 & 82.2 & 51.5 & 24.5	& 78.7	& 31.5	& 21.3	& 47.9	& \cellcolor{magenta!5}\textbf{44.6}	& \cellcolor{green!5}\textbf{51.3} \\
		\midrule
		AUGCO~\cite{prabhu2022augmentation} (Online) & \checkmark & 74.8 & 32.1 & 79.2 & 5.0 & 0.1 & 29.4 & 3.0 & 11.1 & 78.7 & 83.1 & 57.5 & 26.4 & 74.3 & 20.5 & 12.1 & 39.3 & \cellcolor{magenta!5}39.2 & \cellcolor{green!5}45.5 \\
        \rowcolor{aliceblue} C-SFDA (Online) &  \checkmark & 85.9 & 38.1 & 79.2 & 11.9 & 1.1 & 32.0 & 17.1 & 22.9 & 79.7 & 89.4 & 46.6 & 22.0 & 78.4  & 29.6 & 17.4 & 46.0 & \cellcolor{magenta!5}\textbf{43.0} & \cellcolor{green!5}\textbf{49.5} \\
\bottomrule
	\end{tabular}
	}
	\label{table:synthia2city}
	\vspace{-2mm}
\end{table*}
\renewcommand{\arraystretch}{1}
\begin{table}[!h]
	\centering
	\caption{\footnotesize
	Evaluation on \textbf{Cityscapes$\to$Dark-Zurich}. We report mean IoU (mIoU) over 19 common categories between theses datasets.}
	\vspace{-2mm}
	\resizebox{1\linewidth}{!}{
	\begin{tabular}{c|c|c|c|c|c}
		\toprule
		Method  & Source  & TTBN~\cite{nado2020evaluating} &  TENT~\cite{wang2021tent} & AUGCO~\cite{prabhu2022augmentation} &  \cellcolor{aliceblue}C-SFDA (Ours) \\
		\midrule
		mIoU & 28.8 & 28.0 & 26.6  & 32.4 & \cellcolor{aliceblue} \textbf{33.2} \\
        \bottomrule
	\end{tabular}
	}
	\label{table:city2zurich}
	\vspace{-3mm}
\end{table}

\vspace{0.7mm}

\noindent \textbf{Evaluation on Semantic Segmentation Task:}
Table~\ref{table:gta2city} shows the performance on  GTA5$\to$Cityscapes. For this adaptation, we resize the target scenes to $1024\times512$ and use DeepLabV2 for training. We choose this common architecture to be consistent with other recent works. The proposed method outperforms the state-of-the-art SFDA method HCL~\cite{huang2021model} with 19-way averaged mIoU of 48.3\%. Note that, some classes in Cityscapes have very low initial pixel-level accuracy, \eg \emph{Train} category, and it is challenging to obtain satisfactory  performance even with selective pseudo-labeling. It requires mentioning that HCL~\cite{huang2021model} employs historical contrastive loss enforcing additional memory overhead; an undesirable property in most adaptation scenarios. On the other hand, our method utilizes a simple pixel-level prediction reliability measure which is highly computationally efficient and leads to the best mIoU. As regular UDA techniques have the advantage of source data access and most employ highly sophisticated techniques specific to semantic segmentation, they usually perform better than SFDA techniques. However, we find C-SFDA performs comparably to several UDA techniques, \eg CrCDA~\cite{huang2020contextual}. We also evaluate SYNTHIA$\to$Cityscapes (Table~\ref{table:synthia2city}) benchmark, where we use the same DeepLabV2 architecture and adaptation strategy. Compared to the baselines, C-SFDA performs significantly better, with a mIoU improvement of 1.1\% over the previous SOTA. 


\vspace{0.6mm}
\noindent \textbf{Compatibility to Online Adaptation:} As C-SFDA employs batch-wise selection instead of using the whole dataset, it is readily applicable to online fully test-time domain adaptation~\cite{wang2021tent,prabhu2022augmentation}. In contrast to regular SFDA experiments, we only train the model for 1 epoch following prior works~\cite{wang2021tent, prabhu2022augmentation}, without any change to our training settings. In both image classification and semantic segmentation experiments, C-SFDA performs better than previous state-of-the-art methods (Table~\ref{visdaresult} 
-\ref{table:city2zurich}). For instance, we achieve a $3.5\%$ accuracy gain in VisDA image classification (Table~\ref{visdaresult}). In segmentation, we improve the mIoU by $0.8\%$ in Cityscapes$\to$Dark Zurich (Table~\ref{table:city2zurich}) and $4\%$ in  SYNTHIA$\to$Cityscapes (Table~\ref{table:synthia2city}). These gains can be attributed to the adoption of high-quality pseudo-labels right from the beginning of the training. Learning in this manner gives us a head start in producing reliable pseudo-labels for subsequent self-training.
\begin{figure}[htb]
  \centering
  \begin{subfigure}{0.48\linewidth}
    \includegraphics[width=0.9\linewidth]{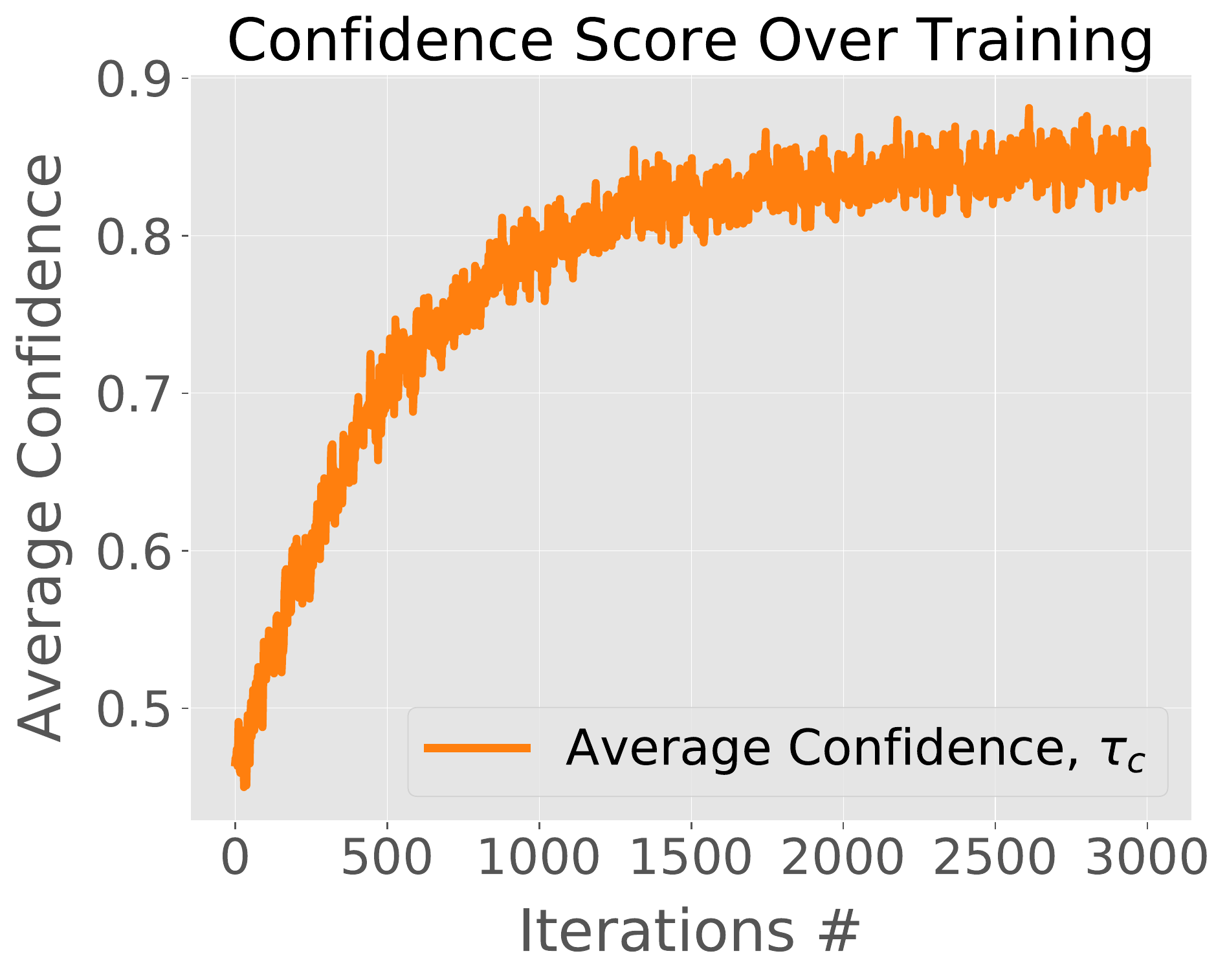}
  \end{subfigure}
  \begin{subfigure}{0.48\linewidth}
    \includegraphics[width=0.9\linewidth]{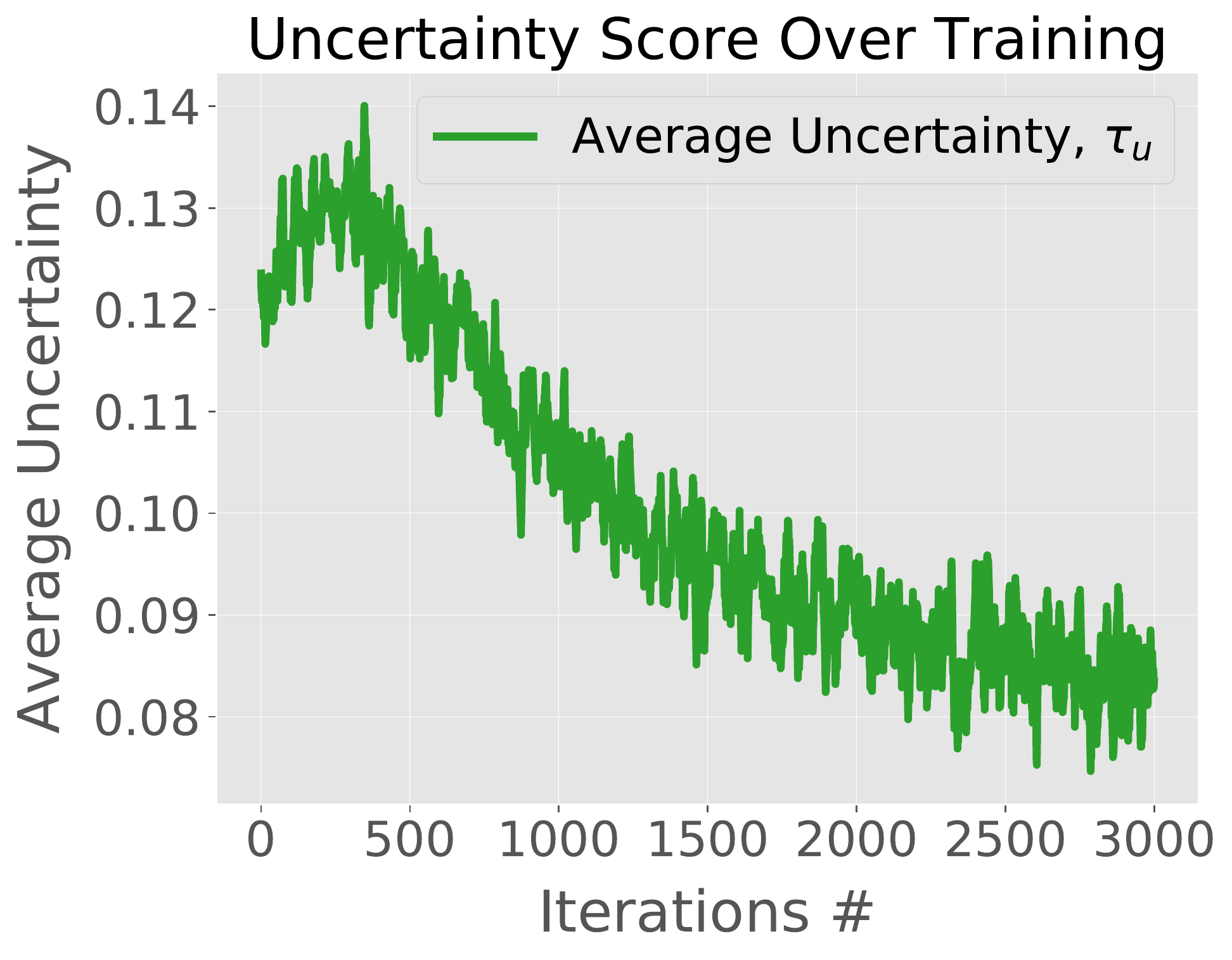}
  \end{subfigure}
  \begin{subfigure}{0.48\linewidth}
    \includegraphics[width=0.9\linewidth]{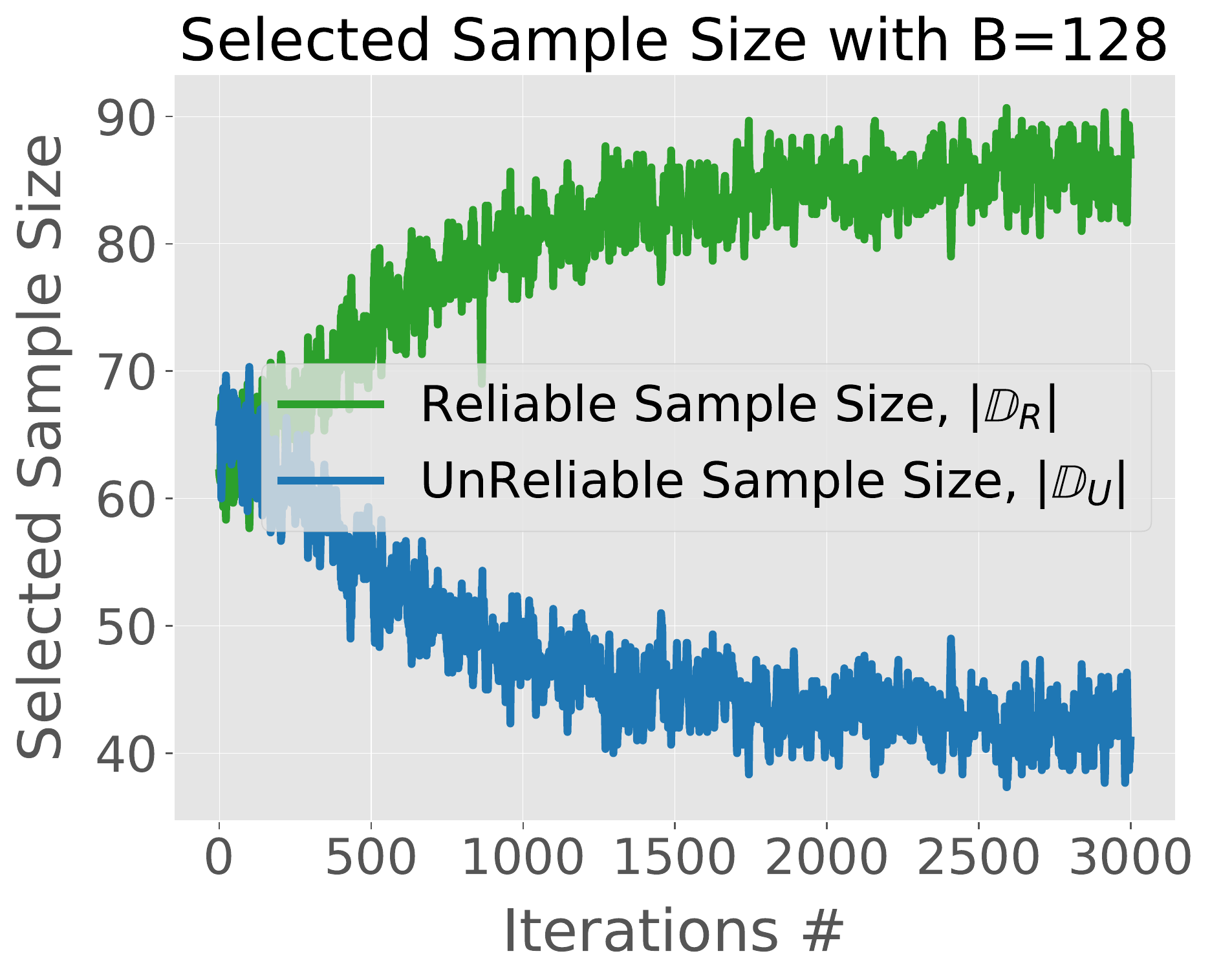}
  \end{subfigure}
  \begin{subfigure}{0.48\linewidth}
    \includegraphics[width=0.9\linewidth]{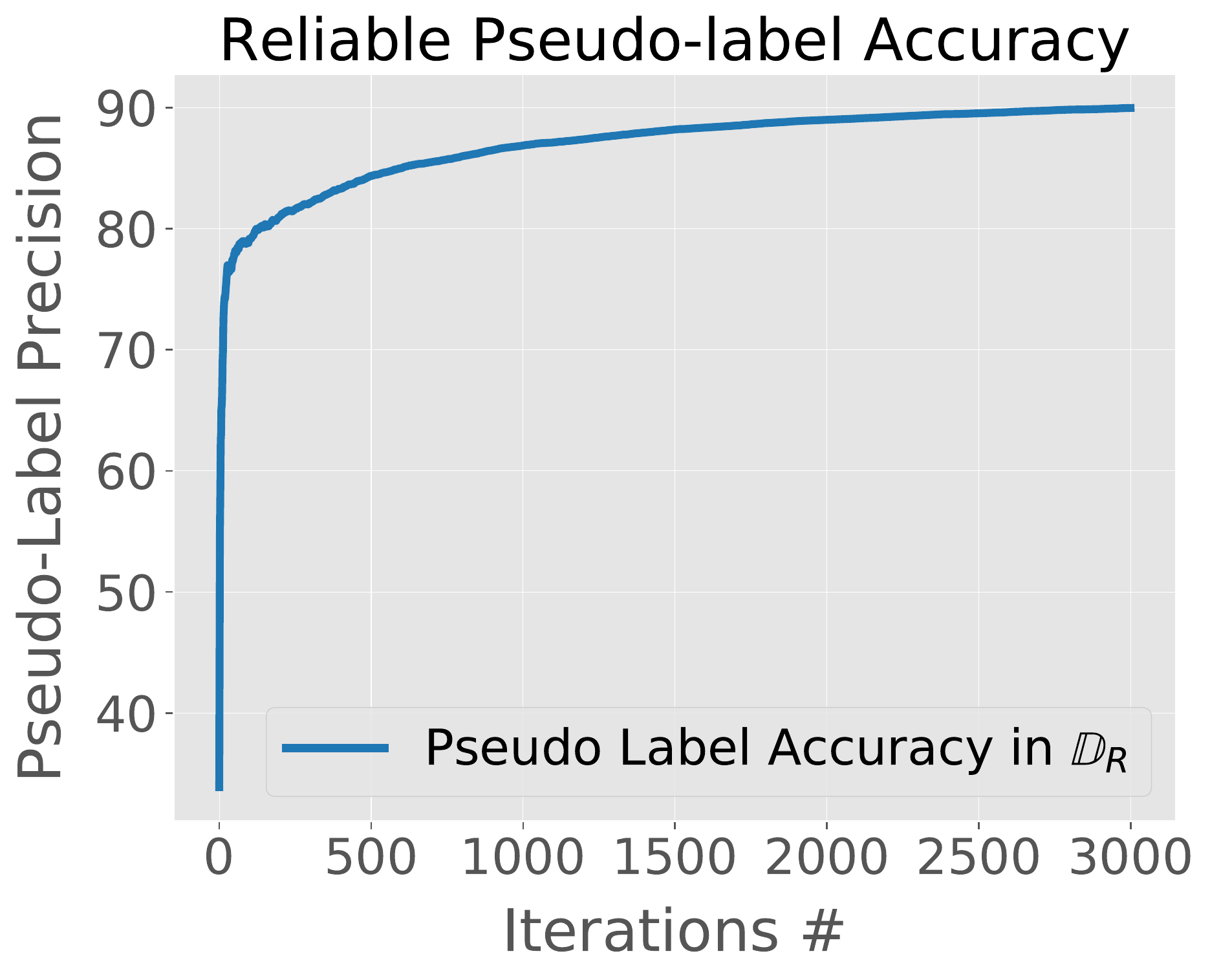}
  \end{subfigure}
  \begin{subfigure}{0.48\linewidth}
    \includegraphics[width=0.9\linewidth]{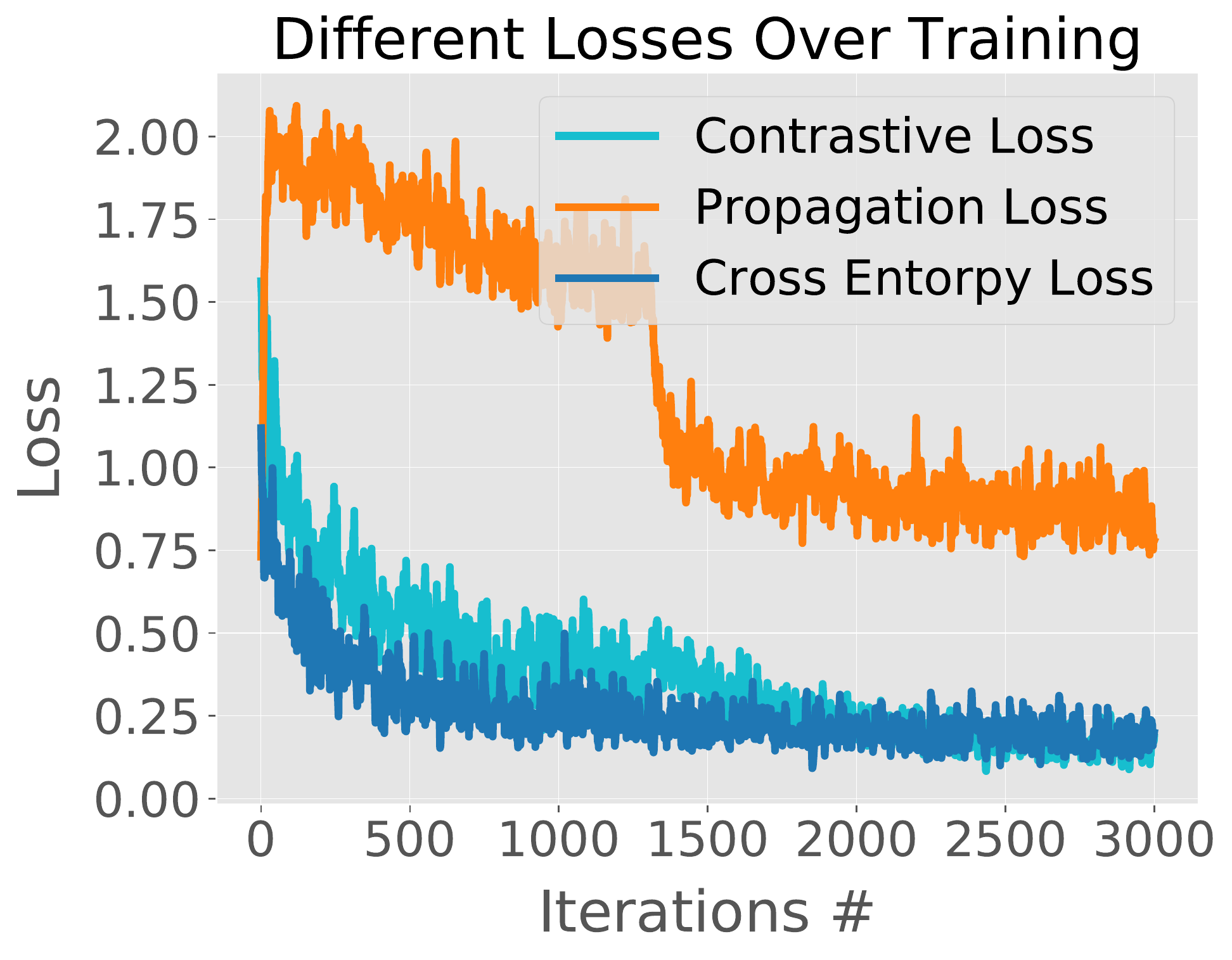}
  \end{subfigure}
  \begin{subfigure}{0.48\linewidth}
    \includegraphics[width=0.9\linewidth]{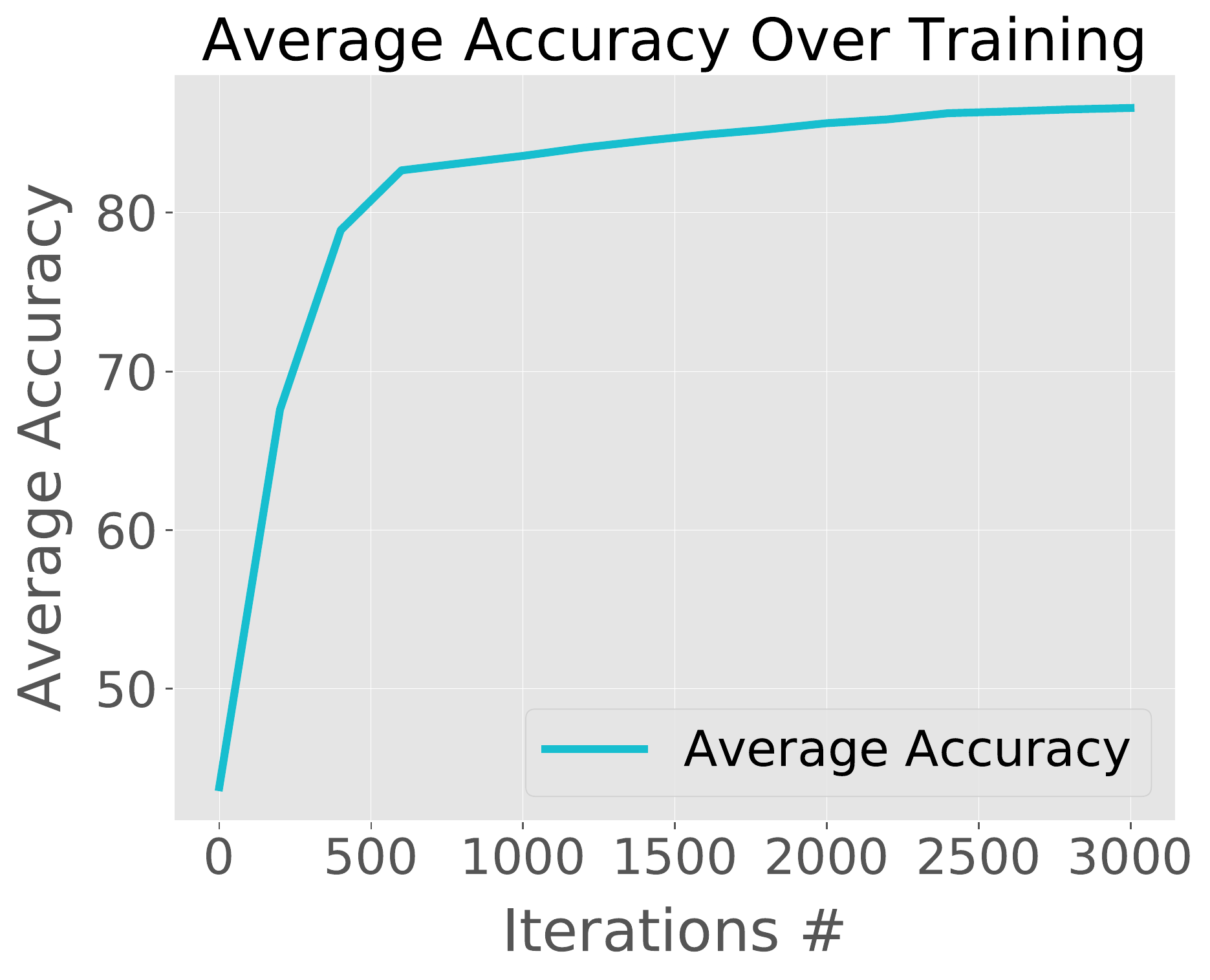}
  \end{subfigure}
  \vspace{-1.5mm}
  \caption{\footnotesize \textbf{Training statistics} for {VisDA Dataset}. As the training progresses, \textbf{(a)} average confidence score increases, \textbf{(b)} average uncertainty score decreases, \textbf{(c)} C-SFDA selects more samples as reliable. \textbf{(d)} Among these selected labels for $\mathbb{D}_R$, most of them are accurate as shown by the pseudo-label accuracy. \textbf{(e)} By putting more weight on the cross-entropy loss, we first learn from $\mathbb{D}_R$ and then learn from $\mathbb{D}_U$ by minimizing propagation loss. Contrastive loss is minimized throughout the training for representation learning.  \textbf{(f)} Average (Avg.) accuracy improves significantly.} 
  \vspace{-2mm}
  \label{fig:sup:train_stats}
\end{figure}

\vspace{-2mm}
\subsubsection{Ablation Study}
\begin{figure}[t]
  \centering
  \begin{subfigure}{0.5\linewidth}
    \includegraphics[width=1\linewidth]{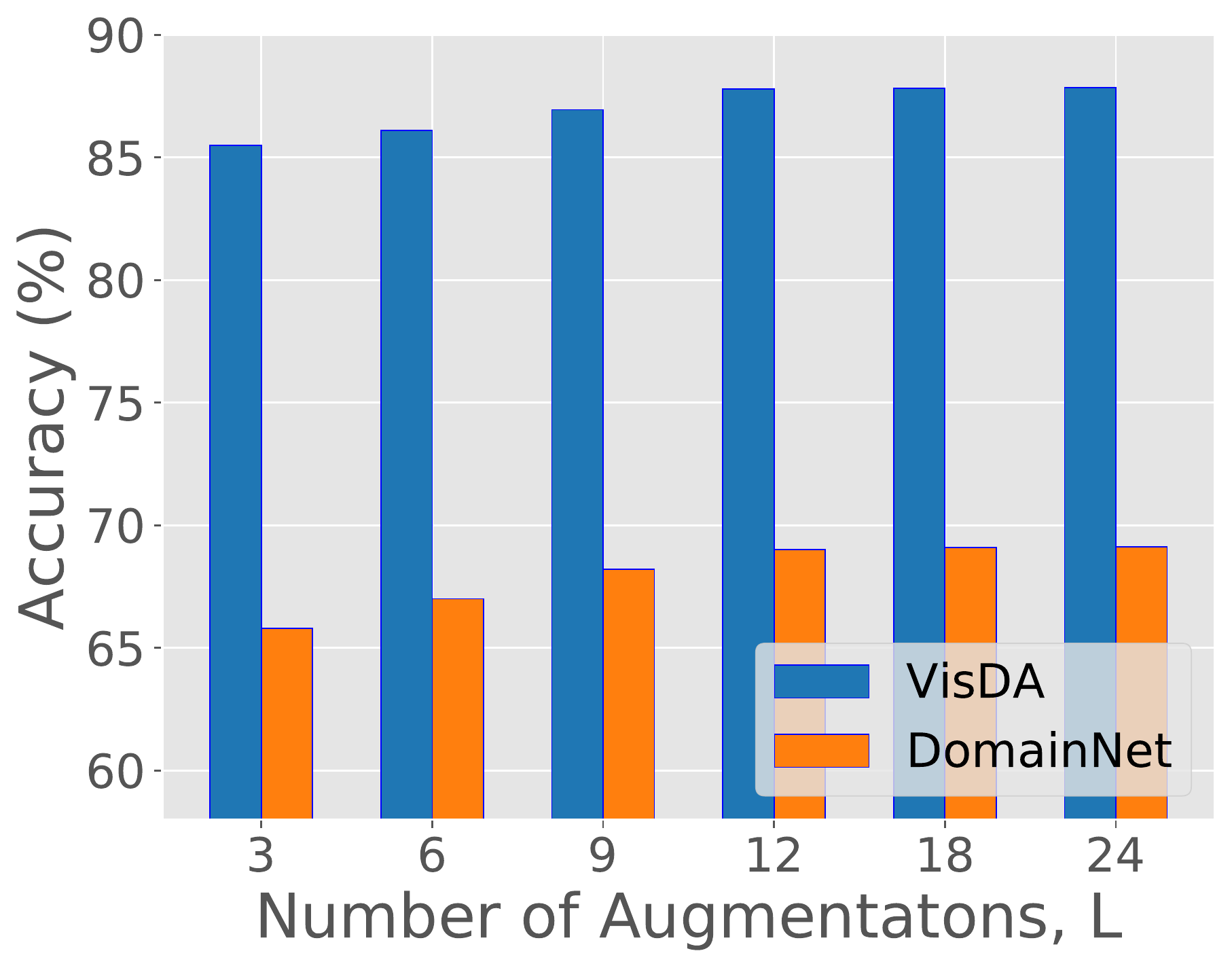}
    \caption{\scriptsize \textbf{Effect of Number of Augmentations}}
    \label{fig:augme}
  \end{subfigure}
  \hfill
  \begin{subfigure}{0.48\linewidth}
    \includegraphics[width=1\linewidth]{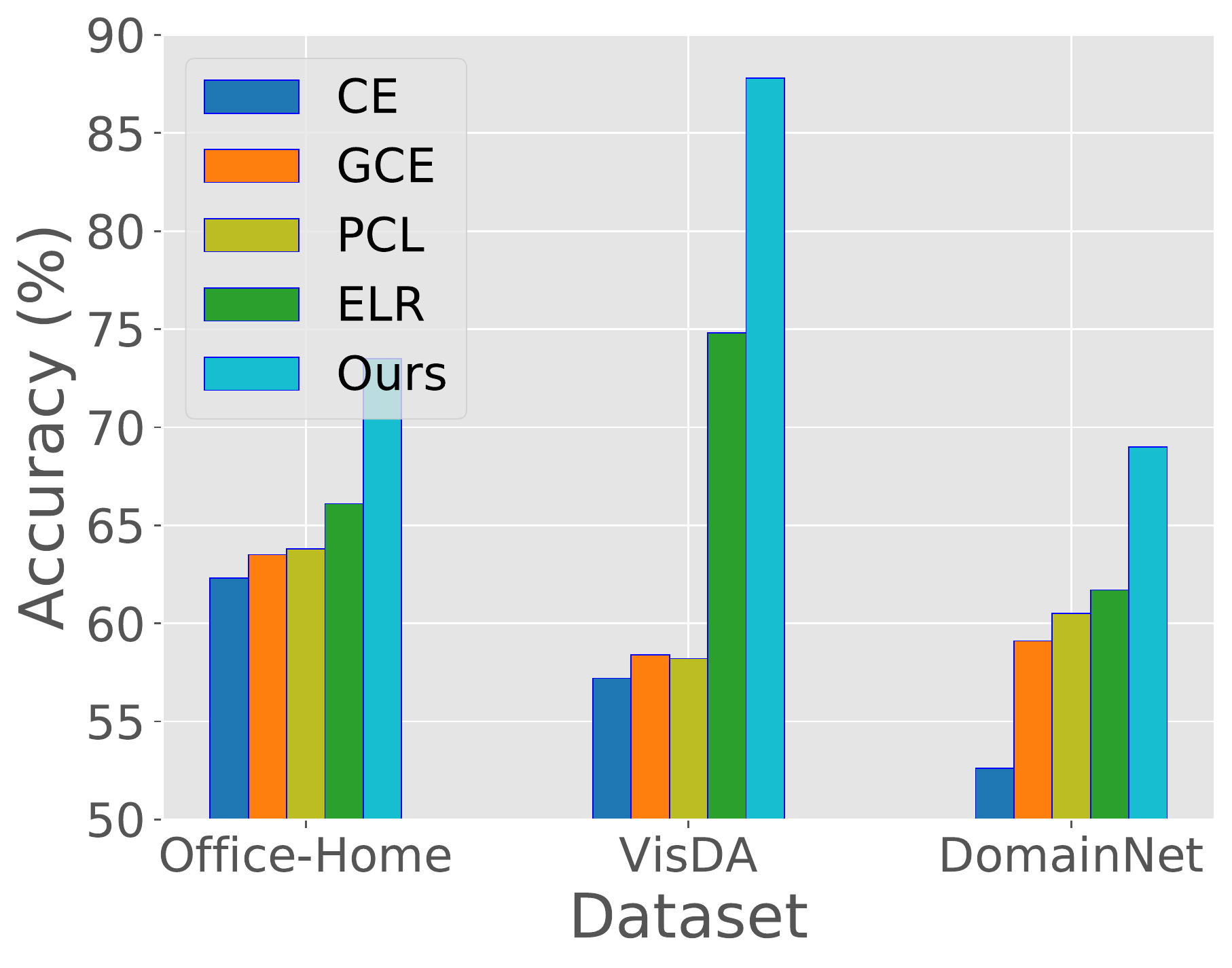}
    \caption{\scriptsize \textbf{LNL Techniques in SFDA}}
    \label{fig:LNL}
  \end{subfigure}
  \vspace{-1.5mm}
  \caption{\footnotesize \textbf{(a)} Ablation with different $L$ shows that we need to consider a sufficient number of augmentations for measuring prediction uncertainty as it plays a crucial role in obtaining SOTA average accuracy. \textbf{(b)} Performance of SOTA noisy label learning methods in SFDA. Due to the presence of unbounded label noise (\ie high noise rates and unknown noise types), traditional LNL struggles to perform well in SFDA settings.}
  \vspace{-3mm}
\end{figure}

\vspace{-1mm}
\noindent \textbf{Does Traditional Label Noise Learning  Help?}
Since our method deals with noisy labels, we explore the literature on label noise learning (LNL) and apply them in the SFDA setting. We consider 3 widely used techniques: GCE~\cite{zhang2018generalized}, PCL~\cite{zhang2021learning}, ELR~\cite{liu2020early} along with regular cross-entropy loss with all pseudo labels and compare their performance on 3 datasets. Fig.~\ref{fig:LNL} shows that traditional LNL techniques may not be suitable for SFDA as they severely underperform compared to C-SFDA. One possible reason could be that SFDA contains unbounded label noise due to an unknown domain shift. In the case of \emph{unbounded label noise}, noise rates are unknown and can be very high; which is in contrast to the general belief of \emph{bounded label noise} where noise rate and type are known priors. In such a scenario, traditional LNL methods struggle to curate label noise. Our method can convincingly perform well in this scenario without any prior knowledge of noise type, rate, \etc. 
\begin{table}[!t]
    \caption{\footnotesize \textbf{Ablation study} with different components of our proposed method. Contrastive learning along with uncertainty plays a vital role in achieving SOTA average accuracy (\%) in image classification benchmarks.}
    \vspace{-2mm}
    \resizebox{1\linewidth}{!}{
    \setlength{\tabcolsep}{2pt}
    \begin{tabular}{ccc|c|ccc|cccc}
        \toprule
           \multicolumn{3}{c}{\centering \bf Selection Strategy} & \multicolumn{1}{c}{\centering \bf Label Bal.} & \multicolumn{3}{c}{\centering \bf Loss} & \multicolumn{4}{c}{\centering \bf Accuracy (\%)} \\
        \cmidrule(l{4pt}r{4pt}){1-3}
        \cmidrule(l{4pt}r{4pt}){4-4}
        \cmidrule(l{4pt}r{4pt}){5-7}
        \cmidrule(l{4pt}r{4pt}){8-11}
          {\centering Conf.} & {\centering Unc.} & {\centering DoC} & {\centering $\lambda_k$} & {\centering $\mathcal{L}_{ce}^R$}& {\centering $\mathcal{L}_\mathcal{P}$} & {\centering $\mathcal{L}_\mathcal{C}$} & {\centering Office-31} & {\centering Office-Home} & {\centering VisDA}  & {\centering DomainNet}  \\ 
        \midrule                 
        \multicolumn{7}{c|}{\centering Self-training ($\mathcal{L}_{ce}$) with all pseudo-labels}  & 81.1 & 62.3 & 57.2 & 52.6 \\
        \midrule  
         \checkmark &  &  & \checkmark & \checkmark & \checkmark &  \checkmark &  87.6 & 69.2 & 85.2 & 65.5 \\  
          \checkmark & \checkmark  & & \checkmark & \checkmark & \checkmark & \checkmark & 89.9 & 71.8 & 87.4 & 68.7 \\  

         \checkmark & \checkmark  & \checkmark & $\times$ & \checkmark & \checkmark & \checkmark & 90.1 & 73.3 & 86.5 & 68.3 \\ 
          \checkmark & \checkmark  & \checkmark  & \checkmark &\checkmark & $\times$ & $\times$ & 88.7 & 71.6 & 85.9 & 67.3 \\
         \checkmark & \checkmark  & \checkmark  & \checkmark & \checkmark & \checkmark & $\times$ & 88.9 & 72.3 & 86.4 & 67.9  \\        
         \midrule
         \rowcolor{aliceblue} \checkmark & \checkmark  & \checkmark & \checkmark & \checkmark & \checkmark & \checkmark & \textbf{90.5} & \textbf{73.5} & \textbf{87.8} & \textbf{69.0} \\      
        \bottomrule
        \end{tabular}}
        \vspace{-2pt}
        \vspace{-5pt}
        \label{tab:ablations}
        \vspace{-1mm}
\end{table}


\vspace{0.6mm}
\noindent \textbf{Effect of Different Selection Criteria:}
Table~\ref{tab:ablations} shows the ablation study with different elements of our proposed method. We first show the performance without curriculum learning where we use all pseudo-labels for fully supervised training with CE loss. Since implementing the curriculum learning requires pseudo-label selection, we analyze the impact of confidence, uncertainty, and DoC here first. It can be observed that each of these metrics can have a significant impact on the overall performance, especially \emph{prediction uncertainty}. As the choice of augmentations plays a vital role in measuring uncertainty, we conduct a detailed study on different augmentation policies (details in supplementary). In Fig.~\ref{fig:augme}, we also analyze the impact of augmentations number $L$ on overall classification performance.  

\vspace{0.6mm}
\noindent \textbf{Effect of Different Loss Functions:}
We also analyze the impact of different loss functions in Table~\ref{tab:ablations}. It can be seen that using only CE loss produces quite satisfactory performance. This indicates learning only from reliable samples is good enough for SFDA. Interestingly, applying propagation loss without contrastive loss may experience performance degradation as we are still using label information. 
This underlines the importance of CL in preventing label noise memorization. 
\begin{wraptable}{r}{3.75cm}
        \centering
        \vspace{-2.5mm}
        \caption{\textbf{\footnotesize Effect of Curriculum}}
        \vspace{-2.5mm}
        \scalebox{0.65}{
        \begin{tabular}{c|c|c}
            \toprule
            \textbf{Dataset} & VisDA-C & DomainNet \\
            \midrule
            \textbf{Curr.} & 87.8 & 69.0 \\
            \textbf{w/o Curr.} & 87.1 & 68.6\\
            \bottomrule
        \end{tabular}}
        \vspace{-2.5mm}
        \label{tab:currculum}
\end{wraptable}
We also show the impact of label balancing here which is only being considered for CE loss. In Figure~\ref{fig:sup:train_stats}, we show some important training statistics of C-SFDA for VisDA dataset. 

\vspace{0.6mm}
\noindent \textbf{Effect of Curriculum:} In Table~\ref{tab:currculum}, we show the impact of curriculum on the performance of our proposed method.

\vspace{-1mm}
\section{Conclusion}
\vspace{-0.5mm}
In this work, we introduce a novel source-free domain adaptation technique exploiting the phenomenon of early memorization of noisy pseudo-labels. Due to the inevitable presence of this phenomenon, we employ a curriculum learning-aided selective self-training strategy that prioritizes learning from highly reliable pseudo-labels and propagating label information to less reliable ones. This leads us to a hyper-parameter independent label selection technique that replaces the need for a label refinement technique. In addition, we utilize contrastive loss-based representation learning that helps generate consistent feature representation and guides the overall adaptation better. Due to the memory-efficient property of our method, C-SFDA can easily be employed for online test-time domain adaptation scenarios. Extensive evaluations show the superior performance of our method on a wide range of image classification and semantic segmentation benchmarks.

\vspace{0.6mm}
\noindent \textbf{Limitations:} We propose to utilize the reliability of generated labels in selective pseudo-labeling. Depending on the domain shift, the initial reliability often varies and can be severely misleading if the domain shift is too large. Such scenarios may require additional measures such as label noise robust self-supervised learning or strongly augmented source domain training. However, we expect our proposed method to be over-restrictive in selecting pseudo-labels whenever such an extreme situation appears. 

{\small
\bibliographystyle{ieee_fullname}
\bibliography{CVPR}
}

\end{document}